\documentclass[letterpaper]{article} 
\usepackage[preprint]{aaai2027}  
\usepackage[hyphens]{url}  
\usepackage{graphicx} 
\usepackage{natbib}  
\usepackage{caption} 
\usepackage{algorithm}
\usepackage{amsmath}
\usepackage{amsfonts}
\usepackage{amssymb}
\usepackage{algpseudocode}
\usepackage{amsthm}
\usepackage[table]{xcolor}
\usepackage{array}
\newcolumntype{G}{>{\color{gray!65}}c}
\usepackage{bbm}

\usepackage{longtable}

\newtheoremstyle{assumptionstyle}
  {6pt}          
  {6pt}          
  {\normalfont}  
  {}             
  {\bfseries}    
  {.}            
  {0.5em}        
  {}             

\theoremstyle{assumptionstyle}

\usepackage{newfloat}
\usepackage{listings}
\DeclareCaptionStyle{ruled}{labelfont=normalfont,labelsep=colon,strut=off} 
\floatstyle{ruled}
\newfloat{listing}{tb}{lst}{}
\floatname{listing}{Listing}

\usepackage{array}
\usepackage{multirow}

\newcolumntype{H}{@{}>{\setbox0=\hbox\bgroup}c<{\egroup}@{}}

\newcommand{\modelname}[1]{\csname #1\endcsname}
\expandafter\def\csname gemini_flash\endcsname{\textsc{Gemini 3 Flash}}
\expandafter\def\csname gemini_flash_35\endcsname{\textsc{Gemini 3.5 Flash}}
\expandafter\def\csname gemini_pro\endcsname{\textsc{Gemini 3.1 Pro}}
\expandafter\def\csname gpt4\endcsname{\textsc{GPT-4.1}}
\expandafter\def\csname gpt5\endcsname{\textsc{GPT-5.4}}
\expandafter\def\csname gpt5_nano\endcsname{\textsc{GPT-5.4 Nano}}
\expandafter\def\csname gpt5_sol\endcsname{\textsc{GPT-5.6 Sol}}
\expandafter\def\csname claude\endcsname{\textsc{Claude Opus 4.8}}

\usepackage{booktabs}

\usepackage{xspace}
\usepackage{capt-of}

\title{Fighting Fire with Fire: On the Feasibility of Protecting\\ Exercises Against AI Cheating}
\author {
  Tobias Braun\equalcontrib,
  Jonas Grebe\equalcontrib,
  Louis Rethfeld\equalcontrib,
  Marcus Rohrbach
}
\affiliations {
  TU Darmstadt \& hessian.AI, Germany
}

\begin{document}

\maketitle

\begin{abstract}
The widespread adoption of generative AI enables students to outsource cognitive effort to increasingly capable assistants, creating an illusion of competence while undermining the independent reasoning that education aims to cultivate. We investigate whether adversarial machine learning can be repurposed to protect educational exercises against such corrosive reliance. Our approach uses multimodal multiple-choice questions whose visual components can be protected with subtle visual perturbations that steer AI solvers toward designated incorrect answers. These responses form a statistical fingerprint: students who blindly copy a solver reproduce the induced answer pattern more frequently than genuine students. We study the feasibility of this paradigm under realistic black-box assistant assumptions using three of the most common state-of-the-art multimodal language models: Anthropic's \textsc{Claude}, Google's \textsc{Gemini}, and OpenAI's \textsc{ChatGPT}. By using accessible surrogate models, we optimize adversarial perturbations that induce consistent response patterns. Those patterns enable principled detection through statistical hypothesis testing. These findings establish both the promise and the limitations of fighting machine-assisted reasoning with the vulnerabilities of the machines themselves.
\end{abstract}

\section{Introduction}

Generative artificial intelligence is fundamentally changing how people access information and solve problems. Multimodal large language models (MLLMs)  have become readily available through conversational interfaces, enabling millions of users to draft text, answer questions, analyze images, and perform increasingly sophisticated reasoning tasks. While these systems offer substantial productivity gains by automating routine work and supporting learning and creativity, their advance is outpacing society's capacity to adapt. We are still struggling to understand and address the consequences of earlier technological shifts, such as social media, including their effects on attention, information consumption, and critical thinking \cite{bayer2020social}. Generative AI presents an even more profound challenge. As these systems become more capable, they raise fundamental questions about when outsourcing thinking improves human productivity and when it begins to undermine the very skills it seeks to augment.

\begin{figure}[t]
    \includegraphics[width=\linewidth]{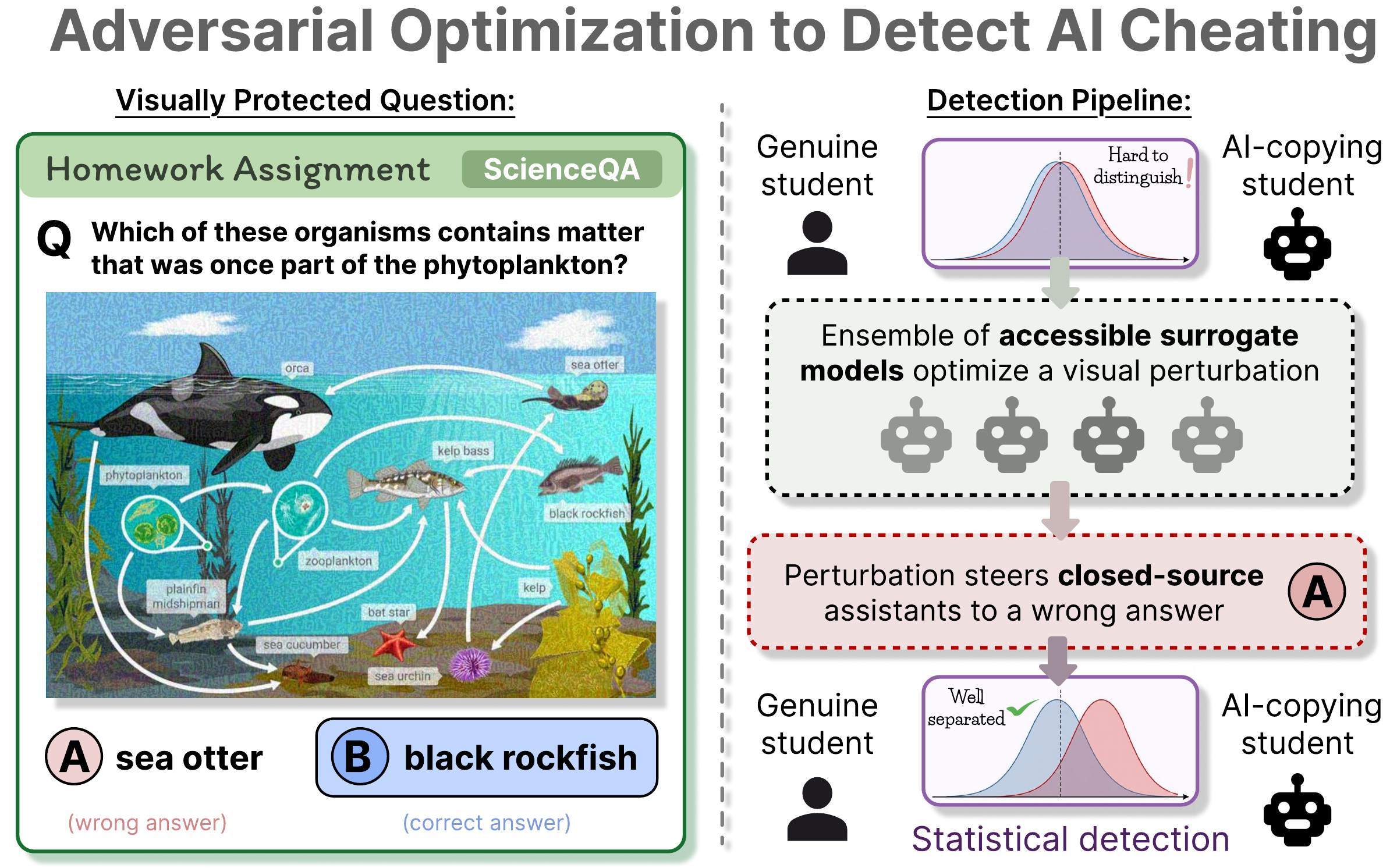}
    \caption{We use an ensemble of surrogate models to optimize subtle visual perturbations that steer closed-source AI assistants toward designated wrong answers. Across an assignment, these controlled errors create a statistical signal that separates genuine responses from blind AI copying.}
    \label{fig:teaser}
\end{figure}

Education lies on the fault line between cognitive augmentation and cognitive erosion: the same tools that can scaffold reasoning may also displace the cognitive effort through which reasoning develops. Learning is not merely about obtaining correct answers, but about the cognitive process required to arrive at them \cite{soderstrom2015learning}. Struggling with challenging problems, making mistakes, and developing perseverance are essential components of acquiring deep understanding and transferable reasoning skills \cite{kapur2014productive}. By directly providing solutions, modern AI systems risk encouraging over-reliance, creating an illusion of competence while bypassing the intellectual effort that produces genuine learning \cite{chirikov2026assessment}. Educational practices developed long before highly capable AI assistants became universally accessible. Today, assignments that were originally designed to foster independent reasoning can often be completed by simply querying a chatbot. Since AI-generated solutions increasingly resemble authentic student work, reliably detecting such misuse is becoming infeasible in practice \cite{goldstein2026student}.

Consequently, educators face an undesirable dilemma. Rather than focusing on teaching, they are increasingly expected to identify AI-generated submissions, an arms race that is both time-consuming and unlikely to succeed as models continue to improve. At the same time, students face strong incentives to use whatever tools maximize short-term academic performance, even if doing so comes at the expense of long-term understanding and intellectual growth.

The broader implications extend beyond education. Frank Herbert's \emph{Dune} imagines a society that responds to dependence on thinking machines by deliberately cultivating human reasoning, memory, and judgment \cite{herbert1965dune}. While fictional, this premise captures an enduring concern: when cognitive tasks are increasingly delegated to machines, education becomes even more essential as a means of cultivating independent, critical-thinking individuals capable of exercising their own judgment. Preserving the integrity of this process is therefore of fundamental societal importance.

In this work, we investigate whether homework assignments can be protected against AI completion by repurposing techniques from adversarial machine learning. Rather than detecting AI-generated solutions after submission, we pursue a preventive strategy: embedding subtle perturbations that steer AI solvers toward specific incorrect answers, preserving the task-relevant semantic content for human students. Across an assignment, the induced errors form a characteristic response pattern that can serve as evidence of blind reliance on a solver. 
Our setting uses multiple-choice questions that include a visual component. This allows efficient grading and provides a controlled answer space in which characteristic error patterns can be detected without analyzing free-form explanations. The visual component supplies the continuous input space required for adversarial optimization.
We examine how well \emph{targeted} perturbations trained on an ensemble of surrogate models transfer to state-of-the-art black-box assistants and whether the resulting answer patterns distinguish genuine students from solver-driven behavior.
Hypothesis testing is used to assess whether a student's responses are more consistent with independent problem solving or blind reliance on an assistant. Figure \ref{fig:teaser} summarizes our approach. 

Overall, our results show that even limited attack transfer to closed-source assistants suffices to build a shared \(20\)-question assignment. Within a single assignment, our method supports detection across multiple frontier \textsc{Claude}, \textsc{Gemini}, and \textsc{GPT} assistants. Under the educator-provided student model and the assumed response structure,
the detector identifies at least \(95\%\) of blind-copying cases while falsely
flagging fewer than \(8\) in \(10{,}000\) genuine students. It also transfers to one assistant not considered during assignment construction. These results establish the feasibility of detecting sustained copying from AI assistants. Beyond individual detections, the increased \emph{perceived risk} of being caught may discourage blind reliance on AI assistants and thereby help protect educational integrity.

\section{Related Work}

\begin{figure}[t]
    \centering
    \includegraphics[width=0.8\linewidth]{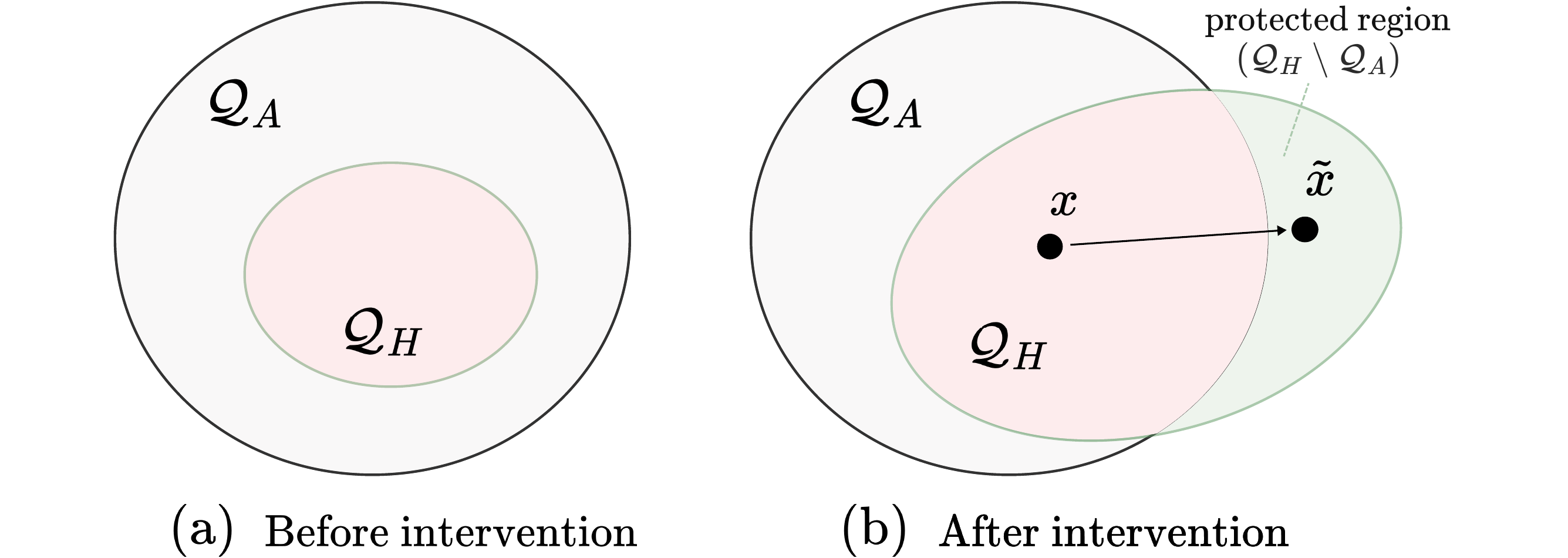}
    \caption{(a) Before intervention, all student-solvable questions lie within the AI-solvable set, $\mathcal{Q}_{\mathrm{H}}\subseteq\mathcal{Q}_{\mathrm{A}}$. (b) Adversarial steering maps $x$ to $\tilde{x}$, creating a protected region $\mathcal{Q}_{\mathrm{H}}\setminus\mathcal{Q}_{\mathrm{A}}$.}

    \label{fig:venn}
\end{figure}

Although the misuse of generative AI in education is recent, academic misconduct has a long history. The term \emph{plagiarius}, used in ancient Rome to describe literary theft, illustrates that concerns over misappropriated intellectual work predate modern educational institutions \cite{mcgillPlagiarism2012}. Accordingly, substantial research has examined how misconduct can be prevented rather than merely detected. Pedagogical approaches seek to reduce students' motivation to cheat by promoting mastery-oriented learning \cite{krou2021achievement}, explicitly teaching academic integrity \cite{perkins2020reducing}, and redesigning assessments to expose the learning process through reflection, and oral verification \cite{lodge2023assessment}. Complementarily, \citet{birksLinking2023} situate the prevention of AI misuse within Situational Crime Prevention \cite{cornishOpportunities2003}, which aims to increase the effort and risk associated with misconduct, reduce its rewards and provocations, and remove excuses. 

Automatic detection offers an appealing response to scalable AI-assisted cheating by distinguishing human-written from model-generated text. Although commercial detectors can perform well under controlled conditions \cite{gehrmann-etal-2019-gltr, waltersEffectiveness2023}, their reliability deteriorates under simple post-processing, stylistic modification, and adversarial adaptation \cite{perkinsSimple2024, nicksLanguage2024}. Such brittleness is particularly problematic in academic settings, where false accusations carry serious consequences and demand a high standard of evidence. OpenAI's withdrawal of its own classifier due to insufficient accuracy further illustrates these limitations \cite{New2024}.

Text watermarking provides a more proactive alternative by embedding detectable signals into generated content \cite{kirchenbauer2024reliability}. However, post-generation watermarking \cite{he2022protecting} depends on responsible deployment and can be bypassed whenever users obtain unmarked outputs, while model-integrated schemes \cite{ICLR2024_a86d17b6} require control over the generator itself. More fundamentally, natural language offers limited capacity for embedding signals that remain detectable under paraphrasing, editing, and translation \cite{chen2024watme}. Although watermarking may strengthen detection, its limited deployment and fragility prevent it from offering a general solution \cite{liu2026position}.

Our work instead builds on the non-robustness of machine-learning systems. Adversarial examples were first described as ``intriguing properties'' of neural networks, arising from geometric properties of learned decision functions that allow small, often imperceptible perturbations to produce large changes in prediction \cite{szegedyIntriguing2013}. Subsequent work distinguished \emph{untargeted} attacks, which divert the model from the correct output without prescribing the resulting prediction, from \emph{targeted} attacks, which steer it toward a predefined adversarial output \cite{papernotLimitations2016,liuDelving2017}. The same phenomenon motivated adversarial training, in which models are exposed to adversarial examples to improve robustness \cite{goodfellowExplaining2015}. Although early work focused primarily on image classifiers, \citet{xuFooling2018} demonstrated attack success rates above 90\% against dense-captioning and visual-question-answering systems. More recent work has extended such attacks to closed-source models \cite{jiaAdversarial2025,hu2026omni}.

These vulnerabilities have also been repurposed defensively. Glaze \cite{shanGlaze2023} protects artistic styles from imitation by image generation models, Nightshade \cite{shanNightshade2024} prevents unauthorized training, and PhotoGuard \cite{photoguard} and EditShield \cite{editshield} disrupt diffusion-based image editing. We extend this defensive use of adversarial non-robustness to education by perturbing multimodal exercises so that AI solvers produce controlled and statistically distinctive errors while the task remains unchanged for human students.
Specifically, we use targeted adversarial perturbations against multimodal large language models to induce an assignment-specific pattern that functions as a statistical watermark for AI misuse and can later be detected through hypothesis testing.
This positions our method primarily within the risk-increasing branch of Situational Crime Prevention: blind reliance on an AI solver becomes more likely to leave verifiable evidence, thereby increasing both the perceived and actual risk of detection. 

\section{Methodology}
\label{sec:methodology}

\begin{figure*}[t]
    \centering
    \includegraphics[width=0.9\linewidth]{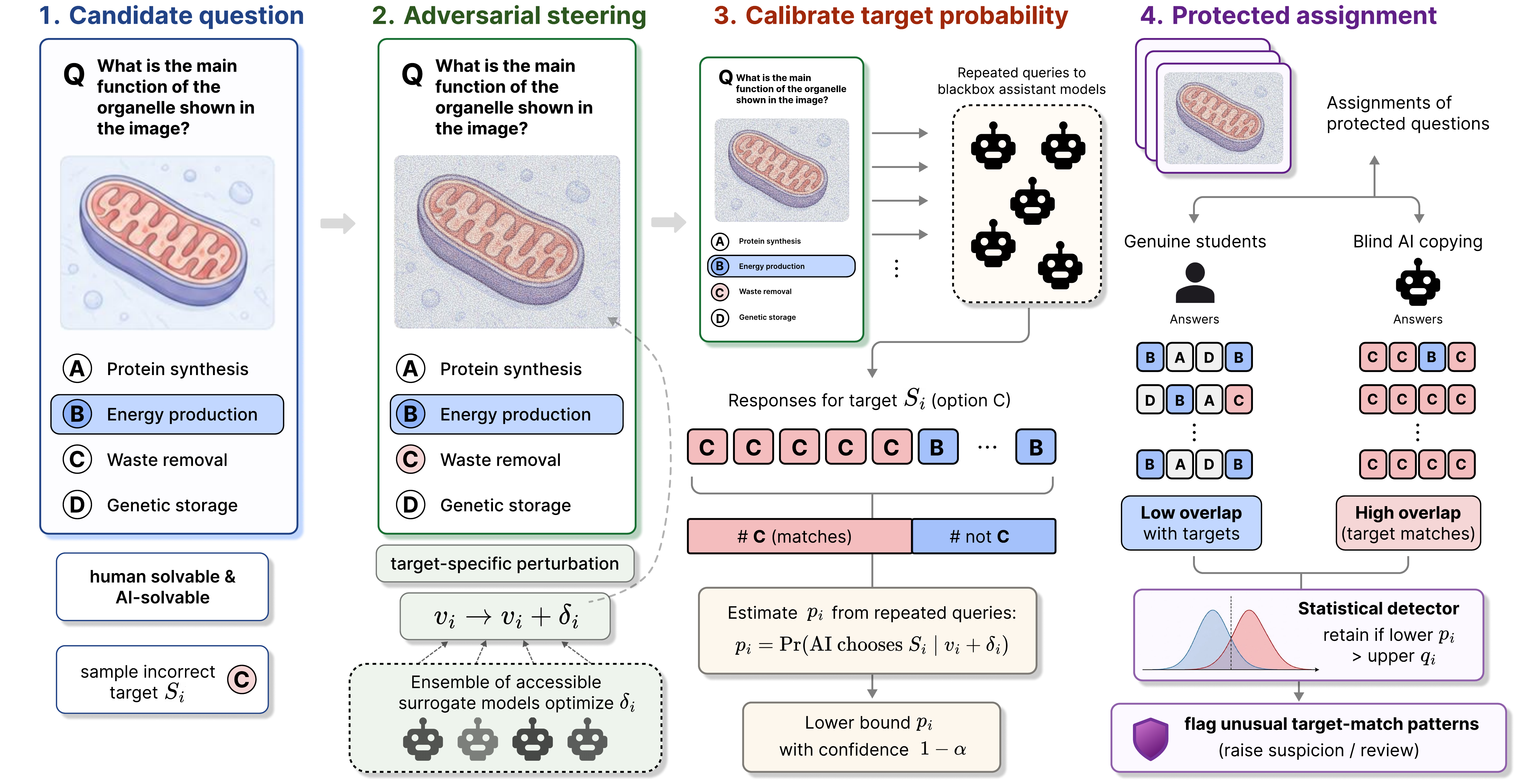}
    \caption{Overview of the proposed framework. First, an incorrect target answer \(S_i\) is sampled for each candidate question (in this example, answer \textbf{C}). Second, an ensemble of surrogate models optimizes a subtle, target-specific perturbation that steers AI solvers toward \(S_i\). Third, repeated queries to candidate black-box assistants estimate the target probability \(p_i\) and a conservative lower bound. Finally, retained questions are assembled into a protected assignment, where unusual overlap between a student's responses and the induced target patterns is evaluated statistically as evidence of blind AI copying.}
    \label{fig:framework}
\end{figure*}

Recent progress on Humanity's Last Exam suggests that AI assistants can solve an increasing share of assessment questions that remain appropriate for human students~\cite{phanBenchmarkExpertlevelAcademic2026b}. Let $\mathcal{Q}_{\mathrm{H}}$ denote the set of questions that students can reasonably solve at the intended level of difficulty, and let $\mathcal{Q}_{\mathrm{A}}$ denote the set of questions that an AI assistant solves reliably. In the limiting case shown in Figure~\ref{fig:venn} (a), we assume \(\mathcal{Q}_{\mathrm{H}}\subseteq\mathcal{Q}_{\mathrm{A}}\), so the questions difficult enough to challenge the assistant are also likely too difficult for the intended students.

Our objective is to alter this relation artificially. Given a question $x\in\mathcal{Q}_{\mathrm{H}}\cap\mathcal{Q}_{\mathrm{A}}$, we construct a perturbed version $\tilde{x}$ that preserves its meaning and intended difficulty for human students but is no longer solved reliably by the assistant:
\[
\tilde{x}\in\mathcal{Q}_{\mathrm{H}},
\qquad
\tilde{x}\notin\mathcal{Q}_{\mathrm{A}}.
\]
This creates the region illustrated in Figure~\ref{fig:venn}(b). Creating this separation requires a modality through which the solver can be misled without altering the task for human students. The visual input provides such a channel. Its high-dimensional redundancy allows small perturbations to remain semantically imperceptible to humans while substantially changing a model's prediction \cite{chen2024watme}. We therefore focus on \textit{multimodal} questions and optimize visually inconspicuous perturbations that increase the difficulty of the task for the AI solver while preserving its meaning and intended difficulty for students. Although lowering the assistant's accuracy would already reduce the expected score of students who blindly copy its answers, our objective is stronger: targeted attacks steer the assistant toward predefined incorrect responses, creating a controlled answer pattern that can be detected statistically across an assessment.

Importantly, we only assume black-box querying access to the AI solver and do \emph{not} require access to weights or model internals. Moreover, our statistical guarantees require neither deterministic assistant failure nor a high overall attack success rate: the adversarial intervention only serves to identify a pool of questions for which the assistant can be steered toward a designated incorrect response with sufficiently high probability, and repeated occurrences of these controlled errors across the final assignment provide the detection signal. For clarity, we first present the detection setup for a single assistant and suppress the assistant index \(j\).

\subsection{Problem Formulation}
\label{subsec:problem_formulation}

We consider an assessment comprising $N$ multimodal multiple-choice questions. Question $i$ contains a visual input $v_i$, a textual component $h_i$, a set of $k_i$ answer options, and a unique correct answer $C_i$. Since a well-prepared, genuine student can also answer all questions correctly, perfect agreement with the correct answers cannot serve as evidence of AI misuse, which motivates our focus on structured error patterns. Let $\mathcal{W}_i$ denote the set of the $k_i-1$ eligible incorrect options. The educator samples a secret target response independently and uniformly,
$
S_i\sim\operatorname{Uniform}(\mathcal{W}_i),
$
and constructs a subtle visual perturbation that is overlaid on $v_i$ to increase the probability that an AI assistant returns $S_i$ when prompted with the questions.

Note that our objective is to detect students who use an AI assistant throughout a substantial part of the assessment and copy its answers without independent verification. Students who use an assistant for only few isolated questions, or who critically inspect and overrule incorrect outputs, fall outside this threat model. 
Let $A_i$ denote the student's submitted response, and define the target-match indicator
\[
Y_i=
\begin{cases}
1, & A_i=S_i,\\
0, & A_i\neq S_i.
\end{cases}
\]
Let
$q_i=P(A_i=S_i\mid H_0)$
and $p_i=P(R_i=S_i\mid\tilde{x}_i),$
where $R_i$ is the response of the deployed assistant to the protected question $\tilde{x}_i$. Under the \textit{genuine-student} hypothesis $H_0$ and the \textit{blind-copying} hypothesis $H_1$, respectively,
\[
Y_i\mid H_0\sim\operatorname{Bernoulli}(q_i),
\qquad
Y_i\mid H_1\sim\operatorname{Bernoulli}(p_i).
\]

\subsection{Statistical Detection}
\label{subsec:statistical_detection}

Given a submitted response sequence, we quantify evidence of blind solver use through the log-likelihood ratio

\begin{equation}
\label{eq:ratio}
L =
\sum_{i=1}^{N}
\left[
Y_i \log \frac{p_i}{q_i}
+
(1-Y_i)\log\frac{1-p_i}{1-q_i}
\right].
\end{equation}

A target match provides strong evidence of blind copying when the evaluated assistant returns $S_i$ frequently, while a genuine student is unlikely to choose the same answer independently. 
The expected contribution of question $i$ to the log-likelihood ratio under blind copying is
$
D_{\mathrm{KL}}\!\left(
\operatorname{Bernoulli}(p_i)
\,\middle\|\,
\operatorname{Bernoulli}(q_i)
\right).
$
In practice, we replace $p_i$ with a calibrated conservative lower bound $\underline{p}_i$ and $q_i$ with a conservative upper bound $\overline{q}_i$. We retain only questions satisfying
$
\underline{p}_i>\overline{q}_i,
$
ensuring that a target match is more likely under blind copying than under genuine behavior. The decision threshold $\tau^\star$ is then chosen as the most stringent attainable threshold satisfying
$P(L\geq\tau^\star\mid H_1)\geq 1-\beta_{\max}$.
Among all thresholds that preserve this required power, this choice minimizes
$P(L\geq\tau^\star\mid H_0)$, which is reported as the resulting Type-I error.
Let $\mathcal{F}=\{L\geq\tau^\star\}$ denote a flag. Given a prior probability \(\pi=P(H_1)\), Bayes' rule yields

\begin{equation}
P(H_1\mid\mathcal F)
=
\frac{
\pi P(L\geq\tau^\star\mid H_1)
}{
\pi P(L\geq\tau^\star\mid H_1)
+
(1-\pi)P(L\geq\tau^\star\mid H_0)
}.
\label{eq:posterior_cheating}
\end{equation}
Thus, a flag can
correspond to a high posterior probability of sustained blind copying only
conditional on the specified genuine-student model and prior. The full multi-assistant fingerprint construction is detailed in
Supp.~\ref{app:multi_assistant}.

\paragraph{Prior specification.}

The prior \(\pi\) denotes the pre-response probability of sustained blind copying from a covered assistant and is specified by the educator based on the assessment context, institutional experience, and available prevalence evidence. ~\citet{chirikov2026assessment} estimate that about 9\% of students who use GenAI engage in measured misconduct. Since our threat model is narrower, we report posterior sensitivity for conservative priors of \(1\%\) and \(5\%\).

\paragraph{Steering and Calibrating $p_i$.}

\begin{figure}[t]
    \centering
    \includegraphics[width=0.8\linewidth]{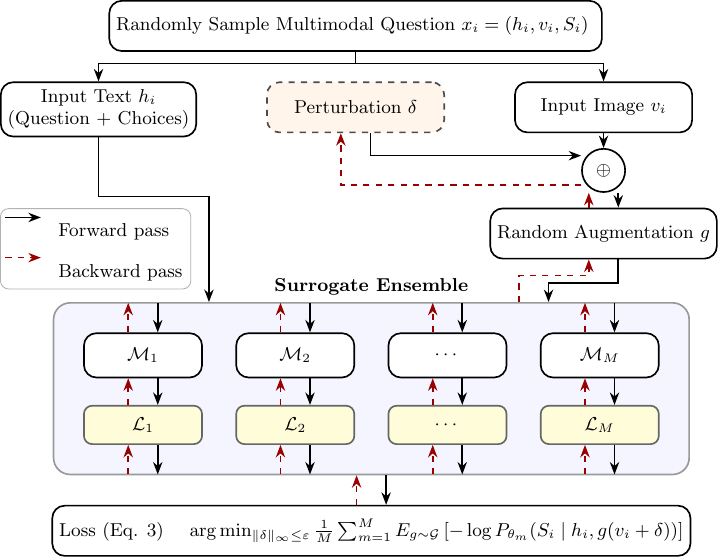}
    \caption{Overview of the perturbation-generation pipeline. Access to a surrogate ensemble enables target-specific optimization that improves transfer to black-box AI assistants.}
    \label{fig:perturbation_pipeline}
\end{figure}

Let $\tilde{v}_i=v_i+\delta_i$ denote the protected visual input, where
$
\|\delta_i\|_\infty\leq\varepsilon
$
bounds the maximum pixel-level change. The textual component $h_i$ contains instructions to answer using only the capital letter associated with the selected option. Steering the assistant toward $S_i$ can therefore be formulated as increasing the probability of a single target token at the first generation step.

Because the deployed assistant is available only through a black-box interface, the perturbation is optimized against an ensemble of $M$ white-box surrogate models. Conceptually, the optimization objective to find a visual perturbation $\delta_i^\star$ is
\begin{equation}
\arg\min_{\|\delta\|_\infty\leq\varepsilon}\frac{1}{M}\sum_{m=1}^{M}\mathbb{E}_{g\sim\mathcal{G}}\left[-\log P_{\theta_m}\!\left(S_i\mid h_i,g(v_i+\delta)\right)\right],
\label{eq:targeted_perturbation}
\end{equation}
where $\theta_m$ denotes surrogate model $m$ and $g\sim\mathcal{G}$ is a randomized image transformation. We use random cropping and resizing so that the perturbation does not overfit to the exact preprocessing pipeline of the surrogate models. 
The objective is optimized using the Momentum Iterative Fast Gradient Sign Method, with the perturbation projected onto the permitted $L_\infty$ region after each update~\cite{dongBoosting2018}.

The surrogate probabilities serve only as optimization signals; after fixing the perturbation, we estimate each target probability \(p_i\) from repeated queries to the deployed assistant and use a conservative lower bound \(\underline{p}_i\) in the detector. To be able to flag various candidate AI assistants, probabilities are calibrated separately and combined via assistant-specific question blocks. Interface-specific estimators and uncertainty propagation are detailed in Supp.~\ref{app:probability_calibration}.

\paragraph{Human-Guided Calibration of $q_i$.}

The item-weighted likelihood ratio (Eq.~\ref{eq:ratio}) requires a model of $q_i$, the probability of a genuine student selecting the steered target $S_i$. We treat educator assessment as the primary source for this quantity, since it allows item difficulty, distractor plausibility, ambiguity, and the intended student population to be judged in context while keeping calibration and item inclusion under human control. The educator, therefore, provides a deliberately conservative upper assessment, $q_i^{\mathrm{H}}$. When representative pilot or historical responses are available, this assessment is supplemented with a one-sided upper confidence bound $q_i^{\mathrm{data}}$ derived from the observed frequency of selecting $S_i$. The operational value is the larger of the two candidates,
$
\overline{q}_i=\max\left\{q_i^{\mathrm{data}},q_i^{\mathrm{H}}\right\}.
$
The educator may increase \(\overline{q}_i\) or exclude the question, but not lower it, as this would weaken the false-positive guarantee. We compare this preferred human-guided model with a data-free fallback that sets \(\overline{q}_i=0.5\) for every target fingerprint; see Supp.~\ref{app:controlled_q_estimates}.

\section{Experimental Setup}
\label{sec:experimental_setup}
Our experiments evaluate whether a pool of ordinary multimodal questions can be converted into a protected assignment with sufficient statistical separation between genuine student behavior and blind assistant copying. The construction follows the four stages in Figure \ref{fig:framework}: (1) Collecting a pool of candidate questions, (2) randomly selecting an incorrect target for each question and optimizing a target-specific perturbation, (3) calibrating the protected target probabilities, and finally, (4) assembling an assignment whose questions satisfy the required detection
power under a specified genuine-student model. 
We evaluate the resulting assignments against black-box AI assistants from three major families: \textsc{Claude Opus 4.8}, two \textsc{Gemini 3} models, and three from the \textsc{GPT-5} family.

\paragraph{Question pool and clean filtering.}

We first select \(100\) image--text multiple-choice questions
each from MMMU~\cite{yueMMMU2024}, ScienceQA~\cite{luLearn2022}, and
MMBench~\cite{liuMMBench2024}. Together, these datasets span expert-level
multidisciplinary reasoning, school-level science, and broad multimodal
understanding. 
To reflect the increasingly realistic threat model in which assistants solve most assessment questions reliably, we reduce the pool of candidate questions per assistant to the questions that the
respective assistant answers correctly before protection. Therefore,
\(N_{\mathrm{correct}}\) differs across assistants and serves
as the denominator for the protected response rates in
Table~\ref{tab:response_shift}. For each remaining question \(i\), we sample one incorrect target
\(S_i\sim\operatorname{Uniform}(\mathcal{W}_i)\). Since \(S_i\neq C_i\), the original
target-response rate is \(0\%\) on every such subset. Protected accuracy (\(\operatorname{Acc}_{\mathrm{prot}}\)) and
target-response rate \((\operatorname{Tgt}_{\mathrm{prot}}\)) are the proportions of this subset answered with
\(C_i\) and \(S_i\), respectively.

\paragraph{Target-specific steering.}

For each question, we initialize $\delta_i=0$ and optimize it toward $S_i$ using the six-model surrogate ensemble described in Table~\ref{tab:white_box_pool}. At each iteration, the perturbed image undergoes randomized cropping and resizing, model-specific differentiable preprocessing, and evaluation by all surrogates. We average their targeted first-token cross-entropy losses and update only $\delta_i$ with MI-FGSM~\cite{dongBoosting2018}, while keeping all model parameters fixed. After each step, the perturbation is projected onto the permitted $L_\infty$ region, and the image is clipped to the valid pixel range. We use $T=50$ attack steps, step size $\alpha=0.5$, and perturbation budget $\epsilon=16$, selected through the sweep in Supp.~\ref{app:attack_hyperparameters}. Figure~\ref{fig:perturbation_pipeline} summarizes our work's steering pipeline; Supp.~\ref{app:educator_q_estimates} and ~\ref{app:semantic_preservation} detail evaluations supporting semantic preservation.

\begin{table}[t] \centering \caption{Ensemble of white-box surrogate models for the perturbation generation. They cover a range of sizes $\sim2B-27B$, different model families and types of vision encoders.} \label{tab:white_box_pool} \resizebox{\linewidth}{!}{ 
\begin{tabular}{lll} \toprule \textbf{VLM} & \textbf{Vision Encoder} & \textbf{Params} \\ 
\midrule Gemma-3-27B-IT \cite{team2024gemma} & SigLIP ViT-So400M & $\sim$27B \\ 
Gemma-4-E4B-IT \cite{team2026gemma} & SigLIP2 ViT & $\sim$4.5B \\ 
SmolVLM2-2.2B-Instruct \cite{marafioti2025smolvlm} & SigLIP ViT & $\sim$2.2B \\ 
Ministral-3-14B-Instruct \cite{liu2026ministral} & SigLIP-based ViT & $\sim$14B \\
InternVL3.5-14B \cite{zhu2025internvl3} & InternViT-6B & $\sim$14B \\ Qwen3-VL-8B-Instruct \cite{bai2025qwen3} & Native ViT + DeepStack & $\sim$8B \\ 
\bottomrule \end{tabular} } \end{table}

\paragraph{Calibrating assistant target probabilities.}

Because API providers expose different levels of output detail, we use
returned answers for provider-independent calibration. For final calibration, we query each selected question--assistant pair
\(K=30\) times under a fixed assistant configuration and
let
\(M_i=\sum_{k=1}^{K}\mathbbm{1}\{R_{i,k}^{\mathrm{prot}}=S_i\}\)
and \(\widehat{p}_i=M_i/K\). Under independent calls,
\(M_i\mid p_i\sim\operatorname{Binomial}(K,p_i)\), from which we compute an
exact one-sided \(95\%\) Clopper--Pearson lower confidence
bound~\cite{clopperUse1934}. We choose \(K=30\) because observing the target
on all \(30\) calls yields
\(\underline{p}_i=0.05^{1/30}\approx0.905\), allowing a lower bound above
\(0.9\). The exact bound and thresholds for alternative budgets
and probability-valued API
outputs are discussed in Supp.~\ref{app:probability_valued_outputs}.

\paragraph{Assembling and evaluating assignments.}
Having calibrated assistant-specific probabilities for individual questions, we next combine them into a shared assignment that supports detection across
multiple assistants. To test whether the resulting assignment also supports
detection of blind copying from assistants not considered at design time, we construct it for the a priori candidate set
$
\mathcal{J}
=
\left\{
\modelname{claude},
\modelname{gemini_flash_35},
\modelname{gpt5_sol}
\right\},
$
and subsequently evaluate it on assistants outside this set, namely \(\modelname{gemini_flash}\), \(\modelname{gpt5}\), and
\(\modelname{gpt5_nano}\).
We consider two models of genuine-student behavior: an educator-guided model and a conservative fallback model. The educator bounds are
obtained through independent assessments by five PhD-level researchers,
taking the maximum estimate as the conservative upper bound. As a conservative data-free alternative, we set
\(\overline{q}_i=0.5\) for every question.  Candidates are ranked by their conservative
informativeness and combined into a shared \(N=20\) assignment, while
requiring at least \(95\%\) detection power for every \(j\in\mathcal{J}\).
Questions selected for multiple assistants are included only once, but
contribute to each fingerprint for which they remain informative. For every
assistant, we choose the most stringent likelihood-ratio threshold that
preserves the required power and report the resulting Type-I error.
Posterior
sensitivity is reported for priors \(\pi\in\{0.01,0.05\}\). Further details are provided in Supp.~\ref{app:controlled_q_estimates}.

\section{Results}
\label{sec:results}

\begin{figure*}[t]
    \centering
    \includegraphics[width=\textwidth]
    {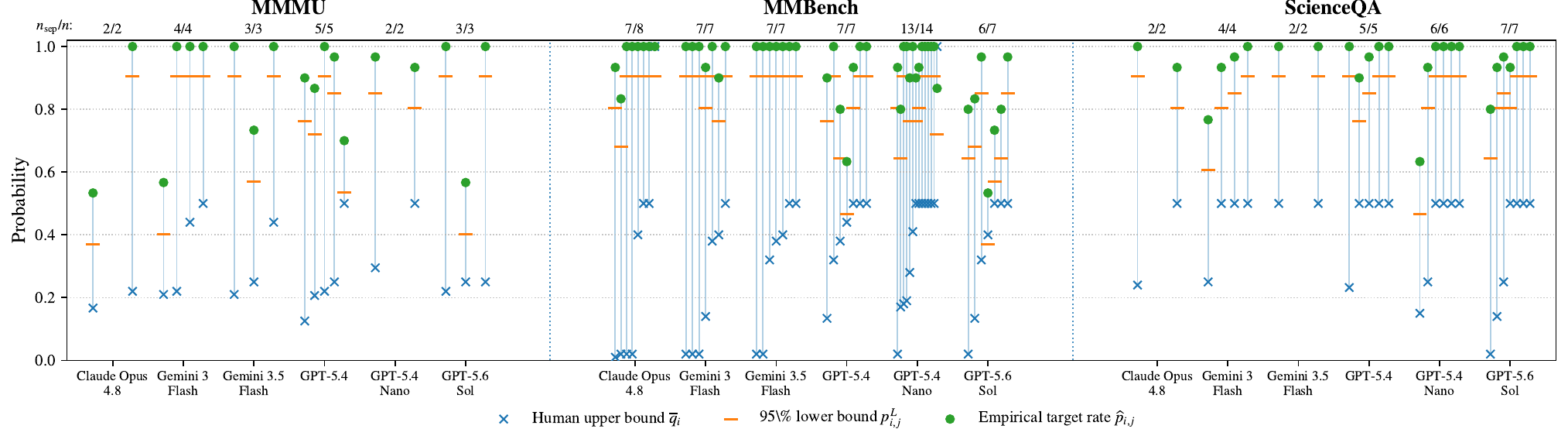}
    \caption{Calibration of assistant-specific target probabilities across
datasets. Each triplet represents one assistant-question pair: circles show
the empirical target rate \(\widehat{p}_{i,j}\) from \(30\) queries,
horizontal markers its one-sided \(95\%\) Clopper--Pearson lower bound
\(\underline{p}_{i,j}\), and crosses the human-established upper bound
\(\overline{q}_i\). Counts report \(n_{\mathrm{sep}}/n\), where
\(n_{\mathrm{sep}}\) denotes pairs satisfying
\(\underline{p}_{i,j}>\overline{q}_i\).}
    \label{fig:target_probability_calibration}
\end{figure*}

\paragraph{Original-to-protected response shifts.}

\begin{figure}
    \centering
    \includegraphics[width=0.9\linewidth]{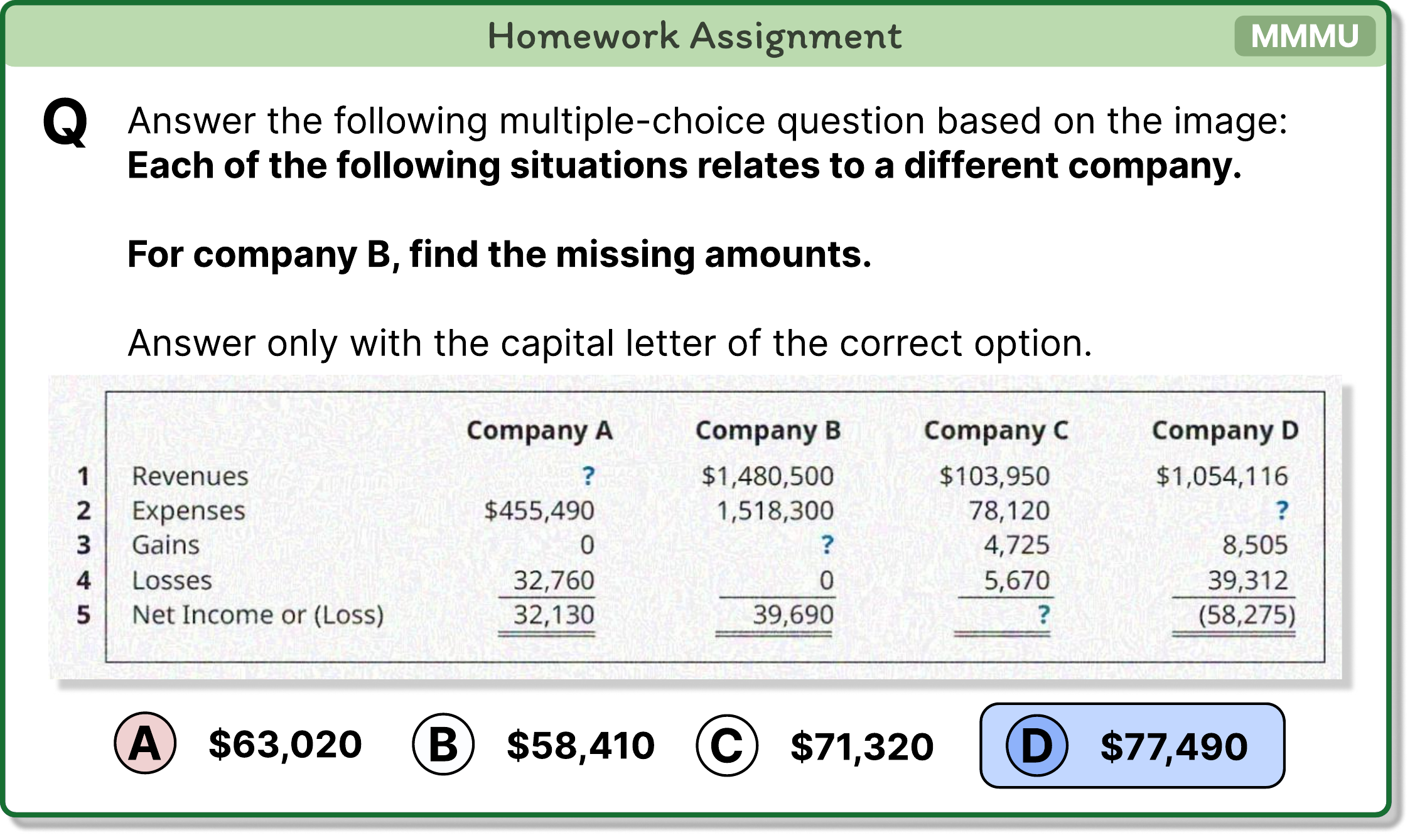}
    \caption{Example protected question from MMMU~\cite{yueMMMU2024}. The overlaid visual perturbation preserves the task-relevant semantics for human students while steering the AI solver from the correct answer (D) toward the designated incorrect target (A). When combined with similarly protected questions in a larger assignment, such controlled errors form a characteristic response pattern that can provide statistical evidence of sustained blind AI copying.}

    \label{fig:qualitative}
\end{figure}

Table~\ref{tab:response_shift} summarizes assistant behavior after protection, and Figure \ref{fig:qualitative} provides a qualitative example of a protected task. We report the protected accuracy, the protected target-response rate, and the non-target error rate,
$
\operatorname{Err}_{\mathrm{other}}
=
100
-
\operatorname{Acc}_{\mathrm{prot}}
-
\operatorname{Tgt}_{\mathrm{prot}}.
$
To distinguish targeted steering from generic disruption, we additionally report
$
\operatorname{TgtShare}
=
\frac{\operatorname{Tgt}_{\mathrm{prot}}}
{100-\operatorname{Acc}_{\mathrm{prot}}}
\times 100,
$
the fraction of induced errors that coincide with the target \(S_i\).

The intervention is most consistently targeted on MMBench, where assistants return \(S_i\) on average in \(8.18\%\) of cases, and \(93.86\%\) of all induced errors coincide with the designated target. This indicates that most induced errors align with the target rather than reflecting generic degradation.
MMMU exhibits a weaker and more heterogeneous response shift, with a mean target rate of \(3.57\%\) and a target share of \(37.00\%\). Here, many induced errors do not reach \(S_i\), which may reflect the greater reasoning complexity of the questions and, in turn, an increase in model uncertainty. ScienceQA is the least steerable dataset overall, with a mean target rate of \(1.87\%\), although \(47.25\%\) of induced errors still match the target. This aligns with prior work that found many ScienceQA questions can be solved largely from the accompanying text~\cite{huang2025mmevalpro}, limiting the influence of perturbations applied only to the image. Therefore, steerability depends not only on the assistant but also on the role of the visual modality and the task structure.

\begin{table}
\centering
\small
\caption{Assistant responses on assistant-specific clean-correct subsets across three datasets. \(N_{\mathrm{correct}}\) serves as the denominator for all protected rates, so subsets may differ across assistants. \(\operatorname{Err}_{\mathrm{other}}\) denotes non-target errors, while \(\operatorname{TgtShare}\) is the share of induced errors matching \(S_i\).}
\label{tab:response_shift}
\setlength{\tabcolsep}{4pt}
\resizebox{\columnwidth}{!}{
\begin{tabular}{llcGcccc}
\toprule
\textbf{Dataset} & \textbf{Assistant}
& \(N_{\mathrm{correct}}\)
& \(\operatorname{Tgt}_{\mathrm{orig}}\)
& \(\operatorname{Acc}_{\mathrm{prot}}\)
& \(\operatorname{Tgt}_{\mathrm{prot}}\)
& \(\operatorname{Err}_{\mathrm{other}}\)
& \(\operatorname{TgtShare}\uparrow\) \\
\midrule
\multirow{7}{*}{MMMU}
& \modelname{claude}         & 63 & 0.00 & 88.89 & 1.59 & 9.52 & 14.31 \\
& \modelname{gemini_flash}    & 74 & 0.00 & 89.19 & 2.70 & 8.11 & 24.98 \\
& \modelname{gemini_flash_35} & 64 & 0.00 & 92.19 & 3.12 & 4.69 & 39.95 \\
& \modelname{gpt5_nano}       & 42 & 0.00 & 90.48 & 4.76 & 4.76 & 50.00 \\
& \modelname{gpt5}            & 64 & 0.00 & 85.94 & 4.69 & 9.37 & 33.36 \\
& \modelname{gpt5_sol}        & 66 & 0.00 & 95.45 & 4.55 & 0.00 & 100.00 \\
\cmidrule(l){2-8}
& \textbf{Mean}               & 62.17 & 0.00 & 90.36 & 3.57 & 6.08 & 37.00 \\
\midrule
\multirow{7}{*}{MMBench}
& \modelname{claude}         & 93 & 0.00 & 89.25 & 8.60 & 2.15 & 80.00 \\
& \modelname{gemini_flash}    & 94 & 0.00 & 91.49 & 7.45 & 1.06 & 87.54 \\
& \modelname{gemini_flash_35} & 94 & 0.00 & 94.68 & 5.32 & 0.00 & 100.00 \\
& \modelname{gpt5_nano}       & 73 & 0.00 & 86.30 & 13.70 & 0.00 & 100.00 \\
& \modelname{gpt5}            & 90 & 0.00 & 93.33 & 6.67 & 0.00 & 100.00 \\
& \modelname{gpt5_sol}        & 95 & 0.00 & 92.63 & 7.37 & 0.00 & 100.00 \\
\cmidrule(l){2-8}
& \textbf{Mean}               & 89.83 & 0.00 & 91.28 & 8.18 & 0.54 & 93.86 \\
\midrule
\multirow{7}{*}{ScienceQA}
& \modelname{claude}         & 91 & 0.00 & 93.41 & 2.20 & 4.39 & 33.38 \\
& \modelname{gemini_flash}    & 90 & 0.00 & 97.78 & 0.00 & 2.22 & 0.00 \\
& \modelname{gemini_flash_35} & 89 & 0.00 & 97.75 & 1.12 & 1.13 & 49.78 \\
& \modelname{gpt5_nano}       & 69 & 0.00 & 97.10 & 1.45 & 1.45 & 50.00 \\
& \modelname{gpt5}            & 87 & 0.00 & 96.55 & 1.15 & 2.30 & 33.33 \\
& \modelname{gpt5_sol}        & 94 & 0.00 & 93.62 & 5.32 & 1.06 & 83.39 \\
\cmidrule(l){2-8}
& \textbf{Mean}               & 86.67 & 0.00 & 96.04 & 1.87 & 2.09 & 47.25 \\
\bottomrule
\end{tabular}
}
\end{table}

\paragraph{Calibration and question retention.}

API providers expose different levels of output detail, ranging from top-\(L\) token log-probabilities to only the generated answer. To obtain a provider-independent estimate, we therefore use the returned answers and query each protected question \(K=30\) times under a fixed assistant configuration. Let
$Z_{i,k}=\mathbbm{1}\{R_{i,k}=S_i\},$
$M_i=\sum_{k=1}^{K}Z_{i,k},$ and 
$\hat{p}_i=\frac{M_i}{K}.$
Under independent calls, \(M_i\mid p_i\sim\operatorname{Binomial}(K,p_i)\). We compute an exact one-sided \(95\%\) Clopper--Pearson lower confidence bound~\cite{clopperUse1934},
\begin{equation}
\underline{p}_i=
\begin{cases}
0, & M_i=0,\\[3pt]
F^{-1}_{\operatorname{Beta}(M_i,K-M_i+1)}(0.05), & M_i>0,
\end{cases}
\label{eq:p_clopper_pearson_lower}
\end{equation}
where \(F^{-1}_{\operatorname{Beta}(a,b)}\) denotes the beta-distribution quantile function.  The evidence provided by various \(K\) and probability-valued API outputs are discussed in Supp.~\ref{app:probability_calibration}.

Figure~\ref{fig:target_probability_calibration} compares the empirical
target-response frequencies with their conservative lower bounds and the
human-established upper bounds \(\overline{q}_i\). Across the six covered
assistants, the calibration data contain \(19\), \(50\), and \(26\)
assistant-question pairs for MMMU, MMBench, and ScienceQA, respectively.
The conservative separation condition
\(\underline{p}_{i,j}>\overline{q}_i\) is retained for all \(19\) MMMU
pairs, \(47\) of \(50\) MMBench pairs, and all \(26\) ScienceQA pairs.
Thus, although only a limited fraction of the initial question pool can be
steered toward a designated target, the target response is typically stable
once a transferable assistant-question pair is identified. The calibration
step, therefore, filters uncertain steering effects while
retaining pairs that provide reliable statistical separation.

\paragraph{Assignment-level detection.}

\begin{table*}
\centering
\small
\setlength{\tabcolsep}{3.5pt}
\renewcommand{\arraystretch}{1.05}
\resizebox{\textwidth}{!}{
\begin{tabular}{lllccccccc}
\toprule
\textbf{Assistant}
&
\textbf{Genuine-student model $q$}
&
\textbf{Covered a priori}
&
\(N\)
&
\(N_j^{\mathrm{inf}}\)
&
\(\tau_j^\star\)
&
\(\widehat{\alpha}_j\)
&
\textbf{Power} ($1-\beta$)
&
\(P(H_{1,j}\mid\mathrm{flag}),\,\pi=0.01\)
&
\(P(H_{1,j}\mid\mathrm{flag}),\,\pi=0.05\)
\\
\midrule

\multirow{2}{*}{\modelname{claude}}
& Educator
& \multirow{2}{*}{Yes}
& \multirow{2}{*}{20}
& \multirow{2}{*}{19}
& 10.163
& \(9.10\times10^{-7}\)
& 0.9502
& 0.99991
& 0.99998
\\
& Target fallback
& & & &
0.295
& \(4.02\times10^{-3}\)
& 0.9508
& 0.70493
& 0.92564
\\
\midrule

\multirow{2}{*}{\modelname{gemini_flash_35}}
& Educator
& \multirow{2}{*}{Yes}
& \multirow{2}{*}{20}
& \multirow{2}{*}{16}
& 4.471
& \(3.74\times10^{-4}\)
& 0.9502
& 0.96252
& 0.99258
\\
& Target fallback
& & & &
-0.230
& \(7.53\times10^{-3}\)
& 0.9509
& 0.56055
& 0.86922
\\
\midrule

\multirow{2}{*}{\modelname{gpt5_sol}}
& Educator
& \multirow{2}{*}{Yes}
& \multirow{2}{*}{20}
& \multirow{2}{*}{16}
& 4.316
& \(3.47\times10^{-4}\)
& 0.9500
& 0.96514
& 0.99312
\\
& Target fallback
& & & &
-0.329
& \(9.07\times10^{-3}\)
& 0.9501
& 0.51412
& 0.84647
\\

\midrule
\multicolumn{10}{c}{\textit{Out-of-set evaluation}} \\
\midrule

\multirow{2}{*}{\modelname{gemini_flash}}
& Educator
& \multirow{2}{*}{No}
& \multirow{2}{*}{20}
& \multirow{2}{*}{16}
& 3.275
& \(1.24\times10^{-3}\)
& 0.9522
& 0.88596
& 0.97589
\\
& Target fallback
& & & &
-0.551
& \(1.35\times10^{-1}\)
& 0.9500
& 0.06637
& 0.27029
\\
\midrule

\multirow{2}{*}{\modelname{gpt5}}
& Educator
& \multirow{2}{*}{No}
& \multirow{2}{*}{20}
& \multirow{2}{*}{14}
& -1.035
& \(2.64\times10^{-1}\)
& 0.9500
& 0.03507
& 0.15923
\\
& Target fallback
& & & &
-1.067
& \(2.83\times10^{-1}\)
& 0.9501
& 0.03278
& 0.15009
\\
\midrule

\multirow{2}{*}{\modelname{gpt5_nano}}
& Educator
& \multirow{2}{*}{No}
& \multirow{2}{*}{20}
& \multirow{2}{*}{14}
& -1.017
& \(2.65\times10^{-1}\)
& 0.9500
& 0.03493
& 0.15868
\\
& Target fallback
& & & &
-1.017
& \(2.65\times10^{-1}\)
& 0.9500
& 0.03493
& 0.15868
\\

\midrule
\multirow{2}{*}{\textsc{Any covered assistant}}
& Educator
& \multirow{2}{*}{\(\mathcal{J}\)}
& \multirow{2}{*}{20}
& \multirow{2}{*}{--}
& model-specific
& \(\leq 7.21\times10^{-4}\)
& \(\geq 0.9500\)
& \(\geq 0.93009\)
& \(\geq 0.98578\)
\\
& Target fallback
& & & &
model-specific
& \(\leq 2.06\times10^{-2}\)
& \(\geq 0.9501\)
& \(\geq 0.31860\)
& \(\geq 0.70804\)
\\

\bottomrule
\end{tabular}
}
\caption{Operating characteristics of the shared \(N=20\) assignment under
the educator and target-fallback models. The fallback sets
\(\overline{q}_i=0.5\) for incorrect-target fingerprints.
Thresholds minimize Type-I error subject to at least \(95\%\) conservative
power. The final rows report Type-I error, minimum power, and
corresponding posterior lower bounds across the three covered assistants.}
\label{tab:detection_results}
\end{table*}

Table~\ref{tab:detection_results} shows that the shared assignment provides strong simultaneous coverage of the three assistants included a priori in \(\mathcal{J}\). Under the educator-provided genuine-student model, the detector achieves at least \(95\%\) power, meaning that it flags at least \(95\%\) of modeled blind-copying cases. Its familywise Type-I error is bounded by \(7.21\times10^{-4}\), so fewer than \(8\) in \(10{,}000\) \emph{genuine} students are expected to be flagged. Given a prior blind-copying probability of \(\pi=0.01\) or \(0.05\), a flag corresponds to a conservative posterior probability of at least \(93.0\%\) or \(98.6\%\), respectively. These values quantify the probability that a flagged submission resulted from blind copying.
The educator estimates are consequential. Replacing the target-answer bounds with the conservative fallback \(\overline{q}_i=0.5\) makes the fingerprint more plausible under genuine behavior, increasing the familywise Type-I bound to \(0.02\), or roughly one false flag in fifty genuine cases. The corresponding posterior lower bounds fall to \(31.7\%\) and \(70.8\%\). The assignment also transfers well to \modelname{gemini_flash}, despite not being constructed for it. In contrast, \modelname{gpt5} and \modelname{gpt5_nano} are poorly separated by the selected fingerprints: retaining \(95\%\) power requires Type-I errors of approximately \(0.26\).

\section{Conclusion \& Future Work}

\label{sec:conclusion}

The central contribution of this work is a shift from classifying submitted work to designing assessments in which sustained blind AI-outsourcing leaves a controlled, auditable trace. By combining target-specific visual perturbations, black-box calibration, and assistant-specific likelihood-ratio tests, we construct an assignment whose target fingerprints cover \textsc{Claude}, \textsc{Gemini}, and \textsc{GPT} assistants. Under the educator-provided genuine-student model, the shared detector achieves at least \(95\%\) power with a modeled false-flag probability below
\(8\) in \(10{,}000\) genuine students. This establishes \emph{feasibility} under a defined threat model.
The guarantees assume that assistants remain stable, responses are conditionally independent, and students copy a covered assistant across a large part of the assignment. Our results emphasize that the protection effectiveness relies on educator judgment, underscoring that assessment integrity remains a human-centered process. Future work should improve steering transfer and explore numerical canaries that may yield more distinctive signals. Ultimately, we hope this work helps preserve the independent reasoning that education is meant to cultivate.

\section*{Acknowledgements}
The research was funded by a LOEWE-Spitzen-Professur (LOEWE/4a//519/\allowbreak{}05.00.002-(0010)/93) and benefited from the Excellence Cluster ``Reasonable AI'' funded by the German Research Foundation (Deutsche Forschungsgemeinschaft, DFG) under Germany's Excellence Strategy, EXC-3057. Additionally, the research was partially funded by an Alexander von Humboldt Professorship in Multimodal Reliable AI, sponsored by the Federal Ministry of Research, Technology and Space (BMFTR). For computational resources, we gratefully acknowledge support from the hessian.AI Service Center, funded by the BMFTR under grant no.~16IS22091, and the hessian.AI Innovation Lab, funded by the Hessian Ministry for Digital Strategy and Innovation under grant no.~S-DIW04/0013/003. Tobias Braun's research visit was supported by the ELSA PhD and Postdoc Mobility Programme within the European Lighthouse on Secure and Safe AI, funded by the European Union under Grant Agreement No.~101070617. Views and opinions expressed are, however, those of the authors only and do not necessarily reflect those of the European Union or the European Commission. Neither the European Union nor the European Commission can be held responsible for them.

\appendix

\bibliography{aaai2027}

\clearpage

\appendix
\setcounter{secnumdepth}{2}
\onecolumn

\begin{center}
    {\LARGE\bfseries Fighting Fire with Fire: On the Feasibility of Protecting\\ Exercises Against AI Cheating\par}
    \vspace{0.5em}
    {\Large Supplementary Material\par}
    \vspace{2em}
\end{center}

\paragraph{Overview.}
This supplement provides the full multi-assistant fingerprint construction, additional statistical details, calibration procedures for assistant and genuine-student response probabilities, experimental implementation details, and an extended discussion of assumptions and limitations. Appendix~\ref{app:multi_assistant} defines the shared-assignment construction. Appendix~\ref{app:statistical_detectors} develops the item-weighted likelihood-ratio detector used in the paper and discusses a secondary randomization-calibrated count detector whose assumptions are not satisfied by our current selection protocol. Appendices~\ref{app:probability_calibration} and~\ref{app:controlled_q_estimates} describe assistant and genuine-student calibration. Appendix~\ref{app:experimental_details} provides additional experimental details, Appendix~\ref{app:additional_examples} presents additional qualitative examples, and Appendix~\ref{app:limitations} summarizes the scope and intended use of the method.

\section{Multi-Assistant Fingerprint Construction}
\label{app:multi_assistant}

A shared assignment may contain questions selected because they are
informative for different assistants. For all reported experiments, the
designated incorrect target \(S_i\) is the fingerprint answer. A question
contributes to the fingerprint of assistant \(j\) only when the protected
assistant returns \(S_i\) more reliably than a genuine student is expected
to select it. Consequently, the same question may be informative for
several assistants, for only one assistant, or for none.

\subsection{Assistant-Specific Target Fingerprints}
\label{app:fingerprint_answers}

For question \(i\) and assistant \(j\), define the protected target
probability
\[
p_{i,j}^{\mathrm{tgt}}
=
P\!\left(R_{i,j}^{\mathrm{prot}}=S_i\right)
\]
and its genuine-student counterpart
\[
q_i^{\mathrm{tgt}}
=
P(A_i=S_i\mid H_0).
\]
Assistant probabilities are represented by conservative lower bounds,
while genuine-student probabilities are represented by conservative upper
bounds. The assistant-specific fingerprint answer is
\begin{equation}
F_{i,j}
=
\begin{cases}
S_i,
&
\underline{p}_{i,j}^{\mathrm{tgt}}
>
\overline{q}_i^{\mathrm{tgt}},
\\[3pt]
\bot,
&
\text{otherwise}.
\end{cases}
\label{eq:supp_fingerprint_answer}
\end{equation}
The symbol \(\bot\) denotes that question \(i\) is uninformative for
assistant \(j\).

For every pair with \(F_{i,j}\neq\bot\), define
\[
p_{i,j}^{\mathrm{fp}}
=
P\!\left(R_{i,j}^{\mathrm{prot}}=F_{i,j}\right),
\qquad
q_{i,j}^{\mathrm{fp}}
=
P(A_i=F_{i,j}\mid H_0),
\]
with conservative bounds
\(\underline{p}_{i,j}^{\mathrm{fp}}\) and
\(\overline{q}_{i,j}^{\mathrm{fp}}\). Under the reported target-only
construction,
\[
p_{i,j}^{\mathrm{fp}}
=
p_{i,j}^{\mathrm{tgt}},
\qquad
q_{i,j}^{\mathrm{fp}}
=
q_i^{\mathrm{tgt}}.
\]
Given a submitted answer \(A_i\), the assistant-specific
fingerprint-match indicator is
\[
Y_{i,j}
=
\mathbbm{1}\{A_i=F_{i,j}\}
=
\mathbbm{1}\{A_i=S_i\}.
\]

Let \(\mathcal A\) denote the shared assignment. A selected question need
not inform every covered assistant. Instead, assistant \(j\) is evaluated
on its informative subset
\[
\mathcal A_j
=
\left\{
i\in\mathcal A:
F_{i,j}\neq\bot
\right\}.
\]
A question selected for several assistants appears only once in the
assignment, but contributes to every target fingerprint for which it is
informative.

\subsection{Assistant-Specific Likelihood Ratios}
\label{app:assistant_likelihood_ratios}

For assistant \(j\), the operational conservative log-likelihood ratio is
\begin{equation}
L_{\mathrm c,j}
=
\sum_{i\in\mathcal A_j}
\left[
Y_{i,j}
\log
\frac{\underline{p}_{i,j}^{\mathrm{fp}}}
{\overline{q}_{i,j}^{\mathrm{fp}}}
+
(1-Y_{i,j})
\log
\frac{1-\underline{p}_{i,j}^{\mathrm{fp}}}
{1-\overline{q}_{i,j}^{\mathrm{fp}}}
\right].
\label{eq:supp_full_fingerprint_llr}
\end{equation}
Under the genuine-student and assistant-specific blind-copying hypotheses,
\[
Y_{i,j}\mid H_0
\sim
\operatorname{Bernoulli}\!\left(q_{i,j}^{\mathrm{fp}}\right),
\qquad
Y_{i,j}\mid H_{1,j}
\sim
\operatorname{Bernoulli}\!\left(p_{i,j}^{\mathrm{fp}}\right).
\]
For an attainable threshold \(\tau\), define the conservative Type-I
probability and power
\begin{align}
a_j(\tau)
&=
P_{\overline{\mathbf q}_j^{\mathrm{fp}}}
\left(
L_{\mathrm c,j}\geq\tau
\mid H_0
\right),
\label{eq:supp_assistant_type_i}
\\
s_j(\tau)
&=
P_{\underline{\mathbf p}_j^{\mathrm{fp}}}
\left(
L_{\mathrm c,j}\geq\tau
\mid H_{1,j}
\right).
\label{eq:supp_assistant_power}
\end{align}
We select the most stringent attainable threshold that retains the required
power,
\begin{equation}
\tau_j^\star
\in
\arg\min_{\tau} a_j(\tau)
\qquad
\text{subject to}
\qquad
s_j(\tau)\geq 1-\beta_{\max}.
\label{eq:supp_assistant_operating_point}
\end{equation}
The reported assistant-specific Type-I error is
\(\alpha_j^\star=a_j(\tau_j^\star)\), and the corresponding conservative
power is \(s_j^\star=s_j(\tau_j^\star)\). Under the conditionally
independent Bernoulli model, these quantities are computed from the exact
weighted-sum distribution. Probabilities at the boundaries of \([0,1]\)
are interpreted by continuity and clipped only when required for
finite-precision computation.

When a submission is flagged if any covered fingerprint crosses its
threshold, the union bound gives
\begin{equation}
P\!\left(
\exists j\in\mathcal J:
L_{\mathrm c,j}\geq\tau_j^\star
\mid H_0
\right)
\leq
\sum_{j\in\mathcal J}\alpha_j^\star.
\label{eq:supp_familywise_type_i}
\end{equation}
This controls the familywise Type-I probability without requiring the
assistant-specific statistics to be mutually independent.

\subsection{Shared-Assignment Assembly}
\label{app:shared_assignment_assembly}

Candidates are ranked separately for each assistant by their conservative
expected contribution
\begin{equation}
d_{i,j}
=
D_{\mathrm{KL}}
\left(
\operatorname{Bernoulli}
\left(\underline{p}_{i,j}^{\mathrm{fp}}\right)
\,\middle\|\,
\operatorname{Bernoulli}
\left(\overline{q}_{i,j}^{\mathrm{fp}}\right)
\right).
\label{eq:supp_candidate_informativeness}
\end{equation}
This score is used only as a ranking heuristic. Because the attainable
likelihood-ratio thresholds are discrete, adding an informative pair does
not necessarily improve the finite-sample operating point. All reported
thresholds, Type-I errors, and powers are therefore recomputed exactly from the completed assignment and do not rely on the ranking score itself.

\subsection{Construction Procedure}
\label{app:construction_algorithm}

Algorithm~\ref{alg:supp_joint_assignment} summarizes the target-only
construction. At each step, the current assistant-specific detectors are
calibrated at the required power, and the assistant with the largest
resulting Type-I error is prioritized. The construction uses development probability bounds. After the assignment
is fixed, the reported operating points are recomputed from the independent
holdout calls described in Supp.~\ref{app:assistant_calibration_scope}.

\begin{algorithm}[t]
\small
\caption{Shared multi-assistant target-fingerprint construction}
\label{alg:supp_joint_assignment}
\begin{algorithmic}[1]
\Require Development-calibrated candidates
\(\widetilde{\mathcal Q}\), assistants \(\mathcal J\), assignment size
\(N\), maximum familywise Type-I error \(\alpha_{\max}\), and maximum
Type-II error \(\beta_{\max}\)
\Ensure Assignment \(\mathcal A\), target fingerprints
\(\{F_{i,j}\}\), and construction-stage operating points
\(\{(\tau_j^\star,\alpha_j^\star,s_j^\star)\}\)

\ForAll{\((i,j)\in\widetilde{\mathcal Q}\times\mathcal J\)}
    \State
    \(F_{i,j}\gets
    \begin{cases}
    S_i,
    &
    \underline p_{i,j}^{\mathrm{tgt}}
    >
    \overline q_i^{\mathrm{tgt}},
    \\[3pt]
    \bot,
    &
    \text{otherwise}
    \end{cases}\)
\EndFor

\State Remove questions with \(F_{i,j}=\bot\) for every \(j\)
\State Rank each informative pair \((i,j)\) by decreasing \(d_{i,j}\)
\State \(\mathcal A\gets\varnothing\)

\ForAll{\(j\in\mathcal J\)}
    \State \(\alpha_j^\star\gets1\) and \(s_j^\star\gets0\)
\EndFor

\While{\(|\mathcal A|<N\)}
    \ForAll{\(j\in\mathcal J\)}
        \State
        \(\mathcal A_j\gets
        \{i\in\mathcal A:F_{i,j}\neq\bot\}\)

        \If{\(\mathcal A_j\neq\varnothing\)}
            \State Select the most stringent \(\tau_j^\star\) satisfying
            \(s_j(\tau_j^\star)\geq1-\beta_{\max}\)
            \State
            \(\alpha_j^\star\gets a_j(\tau_j^\star)\) and
            \(s_j^\star\gets s_j(\tau_j^\star)\)
        \EndIf
    \EndFor

    \State
    \(j^\star\gets
    \arg\max_{j\in\mathcal J}\alpha_j^\star\)

    \If{\(\exists i\in\widetilde{\mathcal Q}\setminus\mathcal A:
    F_{i,j^\star}\neq\bot\)}
        \State Add the highest-ranked such question to \(\mathcal A\)
    \ElsIf{\(\widetilde{\mathcal Q}\setminus\mathcal A\neq\varnothing\)}
        \State Add
        \[
        \arg\max_{i\in\widetilde{\mathcal Q}\setminus\mathcal A}
        \sum_{\substack{j\in\mathcal J\\F_{i,j}\neq\bot}}
        d_{i,j}
        \]
        to \(\mathcal A\)
    \Else
        \State \Return \textsc{Infeasible}
    \EndIf
\EndWhile

\ForAll{\(j\in\mathcal J\)}
    \State
    \(\mathcal A_j\gets
    \{i\in\mathcal A:F_{i,j}\neq\bot\}\)
    \State Select the most stringent \(\tau_j^\star\) satisfying
    \(s_j(\tau_j^\star)\geq1-\beta_{\max}\)
    \State
    \(\alpha_j^\star\gets a_j(\tau_j^\star)\) and
    \(s_j^\star\gets s_j(\tau_j^\star)\)
\EndFor

\If{\(
\min_{j\in\mathcal J}s_j^\star<1-\beta_{\max}
\ \text{or}\
\sum_{j\in\mathcal J}\alpha_j^\star>\alpha_{\max}
\)}
    \State \Return \textsc{Infeasible}
\EndIf

\State \Return
\(\mathcal A,\{F_{i,j}\},
\{(\tau_j^\star,\alpha_j^\star,s_j^\star)\}_{j\in\mathcal J}\)
\end{algorithmic}
\end{algorithm}

\paragraph{Optional correct-answer extension.}
The reported experiments use only the designated target responses
\(S_i\). As a possible extension, consider an assistant-question pair for
which the target channel is uninformative but the protected assistant
continues to return the correct answer \(C_i\) more reliably than a genuine
student. Define
\[
p_{i,j}^{\mathrm{cor}}
=
P\!\left(R_{i,j}^{\mathrm{prot}}=C_i\right),
\qquad
q_i^{\mathrm{cor}}
=
P(A_i=C_i\mid H_0).
\]
When
\[
F_{i,j}=\bot
\qquad\text{and}\qquad
\underline p_{i,j}^{\mathrm{cor}}
>
\overline q_i^{\mathrm{cor}},
\]
one may replace \(F_{i,j}=\bot\) by \(F_{i,j}=C_i\) and apply the same
fingerprint construction and likelihood-ratio analysis.

Each returned-answer calibration call records the complete categorical
response, so the same \(K\) calls used to estimate
\(\underline p_{i,j}^{\mathrm{tgt}}\) also provide the observations needed
to estimate \(\underline p_{i,j}^{\mathrm{cor}}\). Likewise, a
genuine-student model that assigns probabilities to all answer options
already provides \(\overline q_i^{\mathrm{cor}}\). The extension can
therefore be evaluated without additional assistant queries or student
assessments when these full response records are available. It may improve
assistant coverage or the attainable operating point, but this is not
guaranteed. The augmented fingerprints, thresholds, Type-I errors, and
powers must be recomputed exactly. If the extension is selected after
inspecting calibration data, that selection must also be incorporated into
the calibration protocol. We do not use the correct-answer extension in
any reported experiment.

\section{Statistical Detection Details}
\label{app:statistical_detectors}

\subsection{Primary Item-Weighted Likelihood-Ratio Detector}
\label{app:likelihood_ratio_detector}

For a fixed assistant-specific fingerprint, the item-weighted statistic
preserves the identity and evidential value of each question. The ideal
log-likelihood ratio is
\begin{equation}
L_j
=
\sum_{i\in\mathcal A_j}
\left[
Y_{i,j}\log\frac{p_{i,j}^{\mathrm{fp}}}{q_{i,j}^{\mathrm{fp}}}
+
(1-Y_{i,j})
\log\frac{1-p_{i,j}^{\mathrm{fp}}}{1-q_{i,j}^{\mathrm{fp}}}
\right].
\label{eq:supp_ideal_likelihood_ratio}
\end{equation}
The expected contribution of question \(i\) under blind copying is
\[
D_{\mathrm{KL}}
\left(
\operatorname{Bernoulli}(p_{i,j}^{\mathrm{fp}})
\,\middle\|\,
\operatorname{Bernoulli}(q_{i,j}^{\mathrm{fp}})
\right).
\]
A fingerprint match is therefore most informative when it is common under
assistant copying and rare under genuine behavior. Non-matches also carry
information, particularly when the assistant fingerprint probability is
high.

Because the true probabilities are unknown, the deployed test uses the
fixed conservative statistic in
Equation~\ref{eq:supp_full_fingerprint_llr}. Retaining only pairs satisfying
\[
\underline{p}_{i,j}^{\mathrm{fp}}
>
\overline{q}_{i,j}^{\mathrm{fp}}
\]
ensures that a fingerprint match has positive relative weight and that the
statistic is increasing in \(Y_{i,j}\). Under the conditionally independent
Bernoulli model, if the true null probabilities satisfy
\[
q_{i,j}^{\mathrm{fp}}
\leq
\overline{q}_{i,j}^{\mathrm{fp}},
\]
calibrating the upper-tail threshold under
\(\overline{\mathbf q}_j^{\mathrm{fp}}\) is conservative. Likewise, if
\[
p_{i,j}^{\mathrm{fp}}
\geq
\underline{p}_{i,j}^{\mathrm{fp}},
\]
power computed under \(\underline{\mathbf p}_j^{\mathrm{fp}}\) is a
conservative lower-bound calculation for the fixed statistic.

For fully specified simple hypotheses and conditionally independent item
responses, the Neyman--Pearson lemma establishes that a possibly randomized
likelihood-ratio test is most powerful at a fixed Type-I level. When the
assistant and genuine-student probabilities are constant across questions,
the likelihood-ratio statistic is a monotone function of the total number
of matches. Its advantage arises when questions carry unequal evidential
value, as in our calibrated setting.

\subsection{Decision Errors and Posterior Interpretation}
\label{app:posterior_interpretation}

Let
\[
\mathcal F_j
=
\left\{
L_{\mathrm c,j}\geq\tau_j^\star
\right\}
\]
denote a flag for assistant \(j\). The assistant-specific Type-I error,
Type-II error, and power are
\[
\alpha_j^\star
=
P(\mathcal F_j\mid H_0),
\qquad
\beta_j^\star
=
P(\mathcal F_j^{c}\mid H_{1,j}),
\qquad
s_j^\star
=
1-\beta_j^\star
=
P(\mathcal F_j\mid H_{1,j}).
\]
These quantities do not directly equal the probability that a flagged
student copied an assistant. Let
\(\pi_j=P(H_{1,j})\) denote the pre-response probability of sustained blind
copying from assistant \(j\). Bayes' rule gives
\begin{equation}
P(H_{1,j}\mid\mathcal F_j)
=
\frac{
\pi_j P(\mathcal F_j\mid H_{1,j})
}{
\pi_j P(\mathcal F_j\mid H_{1,j})
+
(1-\pi_j)P(\mathcal F_j\mid H_0)
}.
\label{eq:supp_posterior_flag}
\end{equation}
If
\[
P(\mathcal F_j\mid H_0)
\leq
\alpha_j^\star
\qquad
\text{and}
\qquad
P(\mathcal F_j\mid H_{1,j})
\geq
s_j^\star,
\]
then
\begin{equation}
P(H_{1,j}\mid\mathcal F_j)
\geq
\frac{
\pi_j s_j^\star
}{
\pi_j s_j^\star
+
(1-\pi_j)\alpha_j^\star
}.
\label{eq:supp_posterior_lower}
\end{equation}
The prior affects only this posterior interpretation. It does not affect
the test statistic, threshold, Type-I error, or power. In our experiments,
we report sensitivity for \(\pi\in\{0.01,0.05\}\), representing low and
moderate pre-response probabilities of the narrower behavior studied here.
These values are not direct transformations of broader AI-use or misconduct
prevalence estimates~\cite{chirikov2026assessment}.

\subsection{Conditional Independence and Joint Response Structure}
\label{app:independence_assumption}

The exact weighted-sum calculations used for
Equations~\ref{eq:supp_assistant_type_i}
and~\ref{eq:supp_assistant_power} assume that fingerprint-match indicators
are conditionally independent across questions under each hypothesis. This
is a substantive modeling assumption. Genuine-student responses may be
correlated through student ability, fatigue, topic overlap, shared
misconceptions, or test-taking strategy. Assistant responses may likewise
be correlated through shared preprocessing, endpoint state, or systematic
model failures.

Question-level marginal bounds alone do not determine the upper tail of the
joint statistic under arbitrary dependence. The reported operating
characteristics are therefore conditional on the product-Bernoulli working
model. A stronger deployment should validate this assumption using complete
response vectors from the intended cohort. Alternatives include calibrating
the null distribution directly from full-assignment student responses,
fitting a latent-ability or hierarchical response model, grouping related
items into blocks, or using conservative simulation and bootstrap
procedures that preserve observed dependence. Under the alternative,
repeated full-assignment assistant runs can be used to assess cross-item
dependence rather than combining separately calibrated marginals.

The familywise bound in Equation~\ref{eq:supp_familywise_type_i} does not
require independence between the assistant-specific tests. It only requires
that each assistant-specific Type-I bound is valid under the chosen joint
response model.

\paragraph{Assumption Justification.}
The relevant variables are not correctness indicators, but matches with
independently selected, item-specific incorrect targets. General factors
such as student ability, fatigue, or test-taking effort may therefore
correlate errors without producing comparable dependence in the events
\(A_i=S_i\). To inflate the detector statistic, a common factor must
repeatedly direct the response toward the particular secret distractors
selected across several heterogeneous questions. We further absorb
item-specific mechanisms into \(q_i\): plausible distractors, common
misconceptions, and ambiguous questions receive larger bounds or are
excluded. Under \(H_{1,j}\), questions are submitted as separate,
context-free requests under a fixed endpoint configuration, so one
assistant response is not provided as context for the next; persistent
model tendencies are instead represented by the calibrated item-specific
probabilities \(p_{i,j}\). We therefore regard conditional independence as
a defensible approximation for the present feasibility analysis,
rather than as a universal property of student or assistant behavior.

\subsection{Secondary Randomization-Calibrated Count Detector}
\label{app:count_detector}

A simpler target-only alternative counts matches with the randomized
incorrect targets. Let
\[
Y_i^{\mathrm{tgt}}
=
\mathbbm{1}\{A_i=S_i\},
\qquad
T_{\mathrm{tgt}}
=
\sum_{i=1}^{N}Y_i^{\mathrm{tgt}}.
\]
A design-based null can be obtained when an eligible set of incorrect
targets \(\mathcal E_i\) is fixed before the final target is drawn and
\[
S_i
\sim
\operatorname{Uniform}(\mathcal E_i)
\]
independently across questions. If the perturbation does not change
genuine-student behavior, then for any fixed submitted answer \(A_i\),
\begin{equation}
P(Y_i^{\mathrm{tgt}}=1\mid A_i,H_0)
=
\begin{cases}
|\mathcal E_i|^{-1},
&
A_i\in\mathcal E_i,
\\[3pt]
0,
&
A_i\notin\mathcal E_i.
\end{cases}
\label{eq:supp_randomization_null}
\end{equation}
Writing \(\rho_i=|\mathcal E_i|^{-1}\), a conservative threshold is
\[
t_\alpha
=
\min
\left\{
t:
P_{\boldsymbol\rho}
\left(
T_{\mathrm{tgt}}\geq t
\right)
\leq
\alpha
\right\}.
\]
This null does not require students to distribute their errors uniformly.
The randomization is introduced by the educator's target-selection
procedure.

The required target-selection assumptions do not hold for the current
experiments. We sample one target for each question, optimize a perturbation
toward that realized target, calibrate its assistant-specific steering
success, and retain question-assistant pairs based on the resulting
separation. Conditioning retention on the success of the realized target
changes the target distribution among retained questions. Consequently,
\(1/(k_i-1)\) is not a valid item-level null probability for the selected
assignment. The randomization-calibrated count detector is therefore not
used for the reported results.

A future protocol could recover the design-based guarantee by fixing
\(\mathcal E_i\) before the final draw and retaining every randomized target
regardless of steering success, by defining \(\mathcal E_i\) using
development data and drawing a fresh target only afterward, or by recording
the actual target-selection probabilities and incorporating them into the
null. A count detector can also be calibrated from question-specific
\(q_i\) values, but then it inherits the same genuine-student modeling
assumptions as the likelihood-ratio detector while discarding item-level
differences in evidential strength.

\section{Calibration of Assistant Fingerprint Probabilities}
\label{app:probability_calibration}

\subsection{Calibration Estimand}
\label{app:assistant_calibration_estimand}

An assistant endpoint is defined by its exact model version, prompt,
decoding policy, tokenizer behavior, preprocessing pipeline, and relevant
API settings. For question $i$, assistant $j$, and answer
$a\in\{S_i,C_i\}$, define
$p_{i,j}(a)=P(R_{i,j}^{\mathrm{prot}}=a)$.
The reported target-only detector uses $a=S_i$. We state the calibration
procedure for both $S_i$ and $C_i$ because the same returned-answer data can
also support the optional correct-answer extension described in
Appendix~\ref{app:multi_assistant}. After the target, perturbation, prompt,
and endpoint configuration have been fixed, we estimate the corresponding
probability from repeated calls. We write
$\underline p_{i,j}^{\mathrm{tgt}}=\underline p_{i,j}(S_i)$ and
$\underline p_{i,j}^{\mathrm{cor}}=\underline p_{i,j}(C_i)$.

\subsection{Returned-Answer Calibration}
\label{app:returned_answer_calibration}

To obtain a provider-independent estimate, we use the returned answer and
query each protected question $K=30$ times under a fixed configuration. Let
$Z_{i,j,k}(a)=\mathbbm{1}\{R_{i,j,k}^{\mathrm{prot}}=a\}$ and
$M_{i,j}(a)=\sum_{k=1}^{K}Z_{i,j,k}(a)$. Under independent calls from a
stationary endpoint,
$M_{i,j}(a)\mid p_{i,j}(a)\sim
\operatorname{Binomial}(K,p_{i,j}(a))$.
The returned-answer estimator is
$\widehat p_{i,j}(a)=M_{i,j}(a)/K$.

We use the exact one-sided $1-\eta_p$ Clopper--Pearson lower confidence
bound~\cite{clopperUse1934},
\begin{equation}
\underline p_{i,j}(a)
=
\begin{cases}
0,
&
M_{i,j}(a)=0,
\\[3pt]
F^{-1}_{\operatorname{Beta}
\left(
M_{i,j}(a),
K-M_{i,j}(a)+1
\right)}
(\eta_p),
&
M_{i,j}(a)>0.
\end{cases}
\label{eq:supp_p_clopper_pearson_lower}
\end{equation}
In the reported experiments, $\eta_p=0.05$, yielding a one-sided $95\%$
per-pair lower bound.

\begin{table*}[t]
\centering
\small
\caption{Minimum target hits required to establish selected one-sided
$95\%$ Clopper--Pearson lower bounds. A dash indicates that the desired
bound cannot be established with the corresponding number of queries.}
\label{tab:cp_thresholds}
\begin{tabular}{ccccccccc}
\toprule
Desired lower bound
& $K=10$
& $K=15$
& $K=20$
& $K=30$
& $K=60$
& $K=100$
& $K=200$
& $K=300$ \\
\midrule
$\underline{p}_i\geq0.50$
& $9/10$ & $12/15$ & $15/20$ & $20/30$
& $37/60$ & $59/100$ & $113/200$ & $165/300$ \\
$\underline{p}_i\geq0.60$
& $9/10$ & $13/15$ & $17/20$ & $23/30$
& $43/60$ & $69/100$ & $132/200$ & $195/300$ \\
$\underline{p}_i\geq0.70$
& $10/10$ & $14/15$ & $18/20$ & $26/30$
& $49/60$ & $78/100$ & $152/200$ & $224/300$ \\
$\underline{p}_i\geq0.75$
& -- & $15/15$ & $19/20$ & $27/30$
& $51/60$ & $83/100$ & $161/200$ & $238/300$ \\
$\underline{p}_i\geq0.80$
& -- & $15/15$ & $20/20$ & $28/30$
& $54/60$ & $87/100$ & $170/200$ & $252/300$ \\
$\underline{p}_i\geq0.85$
& -- & -- & $20/20$ & $29/30$
& $56/60$ & $92/100$ & $179/200$ & $266/300$ \\
$\underline{p}_i\geq0.90$
& -- & -- & -- & $30/30$
& $59/60$ & $96/100$ & $188/200$ & $279/300$ \\
$\underline{p}_i\geq0.95$
& -- & -- & -- & --
& $60/60$ & $99/100$ & $196/200$ & $292/300$ \\
$\underline{p}_i\geq0.99$
& -- & -- & -- & --
& -- & -- & -- & $300/300$ \\
\bottomrule
\end{tabular}
\end{table*}

\subsection{Probability-Valued API Outputs}
\label{app:probability_valued_outputs}

Some APIs provide first-token log-probabilities in addition to the generated
answer. Let $\ell_{i,j,k,t}$ denote the log-probability reported on query
$k$ for token $t$, and let $\mathcal T_j(a)$ contain all accepted
tokenizations of answer $a$, including relevant leading-space variants.
The reported answer probability is
$p_{i,j,k}^{\mathrm{API}}(a)
=\sum_{t\in\mathcal T_j(a)}\exp(\ell_{i,j,k,t})$.

When the returned answer is sampled from the reported distribution,
$p_{i,j}(a)=\mathbb E[p_{i,j,k}^{\mathrm{API}}(a)]$. Its empirical estimate
is
$\widehat p_{i,j}^{\mathrm{API}}(a)
=K^{-1}\sum_{k=1}^{K}p_{i,j,k}^{\mathrm{API}}(a)$.
Because each observation lies in $[0,1]$, Hoeffding's
inequality~\cite{hoeffdingProbability1963} gives the one-sided lower bound
\begin{equation}
\underline p_{i,j}^{\mathrm{API}}(a)
=
\max
\left\{
0,
\widehat p_{i,j}^{\mathrm{API}}(a)
-
\sqrt{
\frac{\log(1/\eta_p)}{2K}
}
\right\}.
\label{eq:supp_api_hoeffding_lower}
\end{equation}
This bound is conservative and distribution-free under independent calls
from a stationary endpoint.

When only the top-$L$ probabilities are returned, some accepted
tokenizations of the answer may be absent from the displayed list. If the
displayed probabilities refer to the original, non-renormalized
distribution, assigning zero to omitted answer tokens yields a censored
lower observation and preserves a conservative interpretation. If the API
renormalizes over the displayed tokens, uses a decoding policy for which
the reported probabilities do not represent returned-answer frequencies,
or does not expose usable probabilities, we instead use the
returned-answer estimator $\widehat p_{i,j}(a)=M_{i,j}(a)/K$ and the
Clopper--Pearson bound in
Equation~\ref{eq:supp_p_clopper_pearson_lower}.

\begin{figure*}[t]
    \centering
    \includegraphics[width=\textwidth]{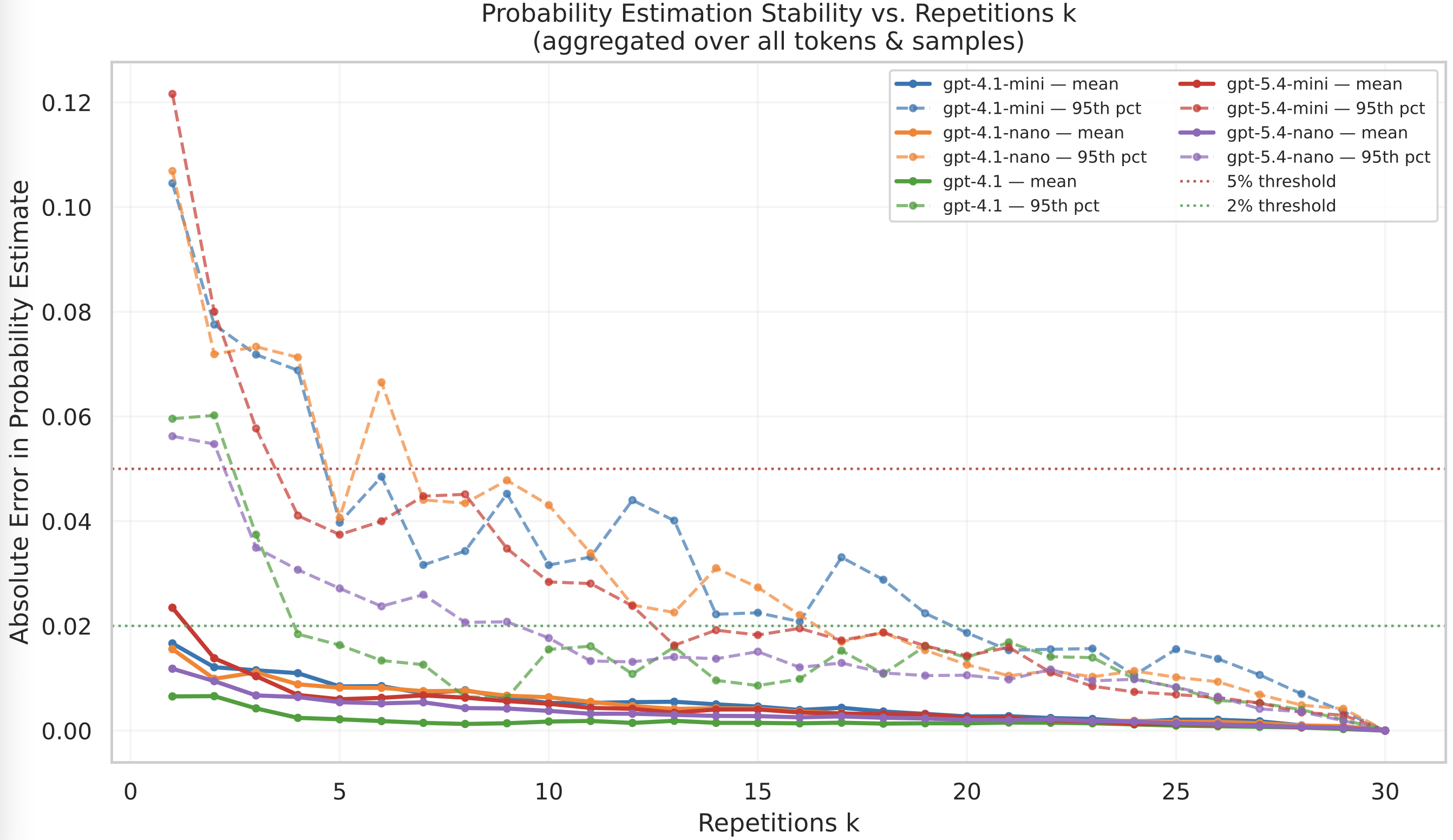}
    \caption{Empirical stability of probability-valued estimates across
    different numbers of repeated API calls. Mean and $95$th-percentile
    absolute deviations from a $30$-call reference are aggregated across
    answer tokens and samples. The diagnostic includes GPT-4.1,
    GPT-4.1 Mini, GPT-4.1 Nano, GPT-5.4 Mini, and GPT-5.4 Nano.
    Horizontal lines mark absolute errors of $0.05$ and $0.02$.
    This analysis is an endpoint-specific convergence diagnostic rather
    than a formal confidence guarantee.}
    \label{fig:probability_stability}
\end{figure*}

Figure~\ref{fig:probability_stability} shows how probability-valued
estimates converge as the number of repeated calls increases. The reported
detector instead uses $K=30$ returned-answer calls for provider-independent
calibration, so these probability-valued estimates do not enter the
reported operating points.

\subsection{Coverage, Holdout Calls, and Model Drift}
\label{app:assistant_calibration_scope}

The bounds above provide per-pair coverage. To obtain simultaneous
calibration coverage $1-\eta_{\mathrm{cal}}$ over a set $\mathcal C$ of
question-assistant-answer triples, compute each bound using an assigned
error probability $\eta_{i,j,a}$ satisfying
$\sum_{(i,j,a)\in\mathcal C}\eta_{i,j,a}\leq\eta_{\mathrm{cal}}$.
The union bound then gives
\[
P\!\left(
p_{i,j}(a)\geq\underline p_{i,j}(a)
\ \text{for every }(i,j,a)\in\mathcal C
\right)
\geq
1-\eta_{\mathrm{cal}}.
\]
If the optional correct-answer extension is evaluated, $\mathcal C$ should
include both $a=S_i$ and $a=C_i$ for every considered
question-assistant pair. Without such an adjustment, assignment-level power
should be interpreted as conditional on the fitted lower-bound model rather
than as a jointly confidence-covered guarantee.

Calls used during attack development, candidate screening, or assignment
construction are not reused for final calibration. After the perturbations
and assignment have been fixed, independent holdout calls are used to
compute the reported lower bounds and operating characteristics. This
separation avoids retaining question-assistant pairs because of favorable
final-calibration noise.

Calibration assumes independent calls from a stationary endpoint. It is
repeated whenever the model version, prompt, decoding policy, tokenizer
behavior, system configuration, or image-processing pipeline changes.
Repeatedly inspecting a fixed-sample interval while collecting additional
observations can invalidate its nominal coverage. Adaptive calibration
therefore requires prespecified batches with an allocated error budget or
an anytime-valid confidence sequence.

\section{Calibration of Genuine-Student Fingerprint Probabilities}
\label{app:controlled_q_estimates}

\subsection{Definition and Controlled Response Data}
\label{app:genuine_student_definition}

For question $i$ and answer $a\in\{S_i,C_i\}$, define
$q_i(a)=P(A_i=a\mid H_0)$. The reported target-only detector uses
$a=S_i$, while $a=C_i$ is included only to support the optional
correct-answer extension. This probability depends on the intended student
population, question difficulty, distractor plausibility, prerequisite
knowledge, ambiguity, and common misconceptions. The detector therefore
requires an upper bound appropriate for the cohort to which it will be
applied.

The preferred source is a set of responses collected under conditions in
which unauthorized assistance can be excluded with high confidence, such
as a proctored examination, supervised classroom assessment, or controlled
pilot study. Let $K_i$ be the number of genuine-student responses and
$G_i(a)=\sum_{k=1}^{K_i}\mathbbm{1}\{A_{i,k}=a\}$. Under the binomial
model, $G_i(a)\sim\operatorname{Binomial}(K_i,q_i(a))$. The exact
one-sided $1-\eta_q$ Clopper--Pearson upper confidence bound is
\begin{equation}
\overline q_i^{\mathrm{data}}(a)
=
\begin{cases}
1,
&
G_i(a)=K_i,
\\[3pt]
F^{-1}_{\operatorname{Beta}
\left(
G_i(a)+1,
K_i-G_i(a)
\right)}
(1-\eta_q),
&
G_i(a)<K_i.
\end{cases}
\label{eq:supp_q_clopper_pearson_upper}
\end{equation}
We use $\eta_q=0.05$ for a one-sided $95\%$ marginal upper bound.

\begin{table}[t]
\centering
\small
\caption{Maximum observed fingerprint selections compatible with selected
one-sided $95\%$ Clopper--Pearson upper bounds. A dash indicates that the
desired bound cannot be established with the corresponding sample size,
even when no participant selects the fingerprint answer.}
\label{tab:supp_cp_upper_thresholds}
\begin{tabular}{ccccc}
\toprule
Desired upper bound & $K=10$ & $K=20$ & $K=30$ & $K=60$ \\
\midrule
$\overline{q}\leq0.50$ & $1/10$ & $5/20$ & $10/30$ & $23/60$ \\
$\overline{q}\leq0.40$ & $1/10$ & $3/20$ & $7/30$  & $17/60$ \\
$\overline{q}\leq0.30$ & $0/10$ & $2/20$ & $4/30$  & $11/60$ \\
$\overline{q}\leq0.25$ & --     & $1/20$ & $3/30$  & $9/60$  \\
$\overline{q}\leq0.20$ & --     & $0/20$ & $2/30$  & $6/60$  \\
$\overline{q}\leq0.15$ & --     & $0/20$ & $1/30$  & $4/60$  \\
$\overline{q}\leq0.10$ & --     & --     & $0/30$  & $1/60$  \\
$\overline{q}\leq0.05$ & --     & --     & --      & $0/60$  \\
\bottomrule
\end{tabular}
\end{table}

The candidate fingerprint answer should be fixed before inspecting the
controlled response frequencies. Selecting a target because it happened to
receive few responses would produce an optimistically biased bound.
Ideally, participants answer the protected version so that its effect on
genuine-student responses is assessed directly. When responses to the
original version are reused, the analysis assumes that the perturbation
does not materially alter genuine-student behavior.

\subsection{Educator-Provided Conservative Assessments}
\label{app:educator_q_estimates}

When representative response data are unavailable, educators may provide a
deliberately conservative upper assessment
$\overline q_i^{\mathrm H}(a)$. The assessment should reflect the intended
student cohort, expected question difficulty, ambiguity, distractor
plausibility, similarity to the correct answer, and whether the response
represents a common misconception. An unusually attractive distractor
receives a correspondingly larger probability or the question is excluded.

In our experiments, five PhD-level researchers independently assessed every
question. All assessors had at least one year of teaching experience and
were explicitly instructed to estimate the response distribution of the
intended student cohort, rather than their own uncertainty or likely
personal response.

\begin{figure}[t]
    \centering
    \includegraphics[width=\linewidth]{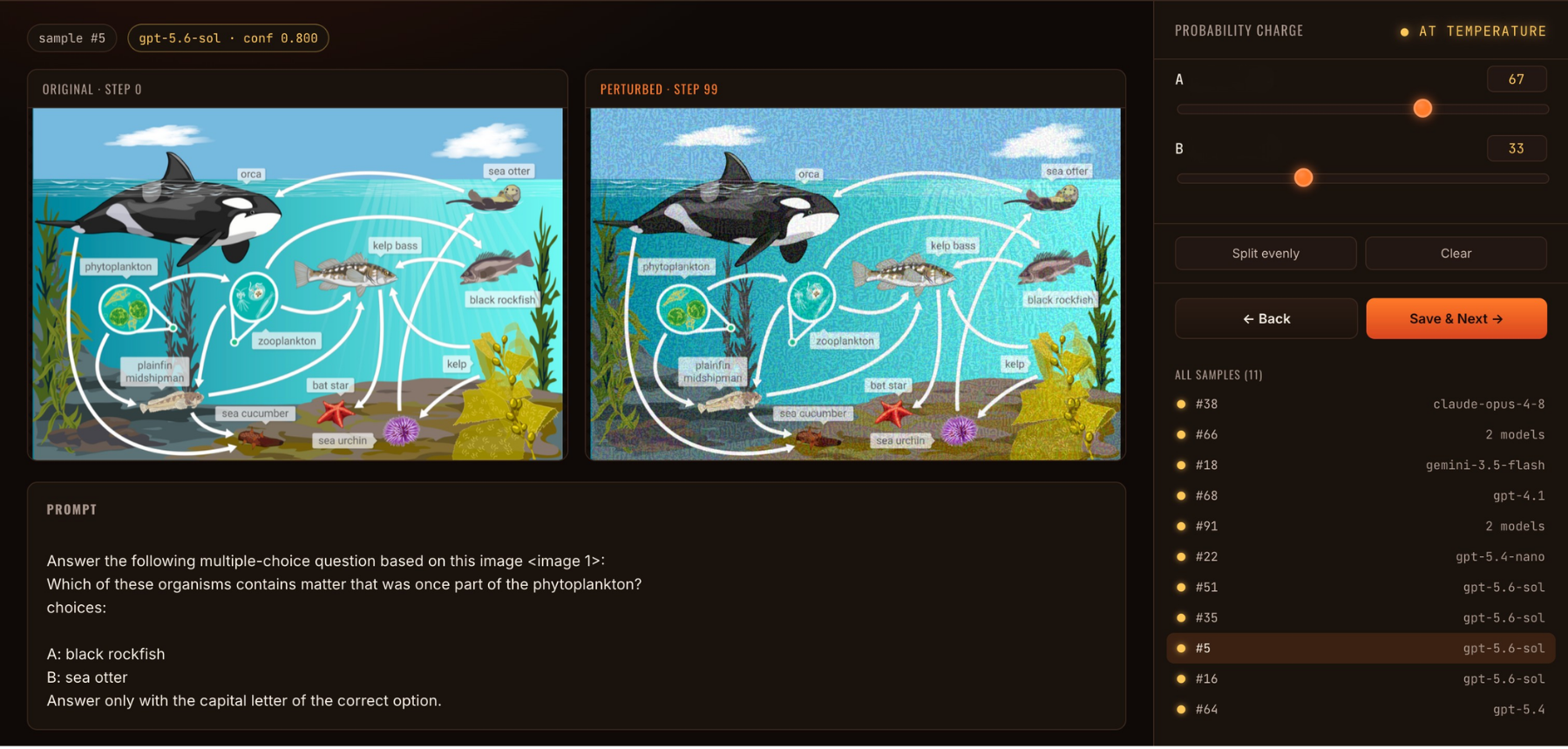}
    \caption{Representative interface used to elicit genuine-student
    response estimates. Five PhD-level researchers independently estimated
    the response distribution of the intended student cohort. Each assessor
    saw exactly one image per question, either clean or protected, while the
    image from the other condition was blacked out. Both versions are shown
    here solely to illustrate the two experimental conditions. Assessors
    were blinded to the assigned condition and adversarial target.}
    \label{fig:educator_interface}
\end{figure}

For each question, every assessor was randomly assigned either the clean or
protected version and never saw both versions of the same item. Assignments
were balanced across assessors and questions such that every item was
evaluated under both conditions. The image corresponding to the unassigned
condition was blacked out in the elicitation interface, although
Figure~\ref{fig:educator_interface} displays both versions for illustration.
Assessors were blinded to which condition they received, the adversarial
target, and the estimates of the other assessors. They assigned a
probability to every answer option, and the target-specific value was
extracted only after elicitation as the probability assigned to $S_i$.

To assess whether protection materially changed the elicited target-response
probability, we evaluated equivalence using a prespecified absolute margin
of $\Delta=0.05$. The \(90\%\) confidence interval was computed using a paired
question-level \(t\)-interval over the clean--protected differences. The mean
difference was $\widehat d=0.0275$, with a \(90\%\) confidence interval of
$[0.0158,0.0392]$. Thus, protected questions received slightly higher
target-response estimates on average, but the entire interval remained
within the equivalence region $[-0.05,0.05]$, meeting the equivalence
criterion under a five-percentage-point margin.

Let $q_{i,e}^{\mathrm H}(a)$ denote assessor $e$'s estimate for answer $a$.
For operational calibration, we conservatively aggregate the assessments as
$\overline q_i^{\mathrm H}(a)
=\max_{e\in\{1,\ldots,5\}}q_{i,e}^{\mathrm H}(a)$.
Because the maximum is taken across assessors and assigned conditions,
assessor disagreement or a condition-specific increase can only raise the
genuine-student bound. This weakens, rather than artificially strengthens,
the resulting evidence for blind copying.

If both controlled response data and educator estimates are available, the
operational bound is
$\overline q_i(a)=
\max\{\overline q_i^{\mathrm{data}}(a),
\overline q_i^{\mathrm H}(a)\}$.
If only one source is available, its corresponding bound is used. We write
$\overline q_i^{\mathrm{tgt}}=\overline q_i(S_i)$ and, for the optional
correct-answer extension,
$\overline q_i^{\mathrm{cor}}=\overline q_i(C_i)$.

Educator-provided values are modeling assumptions rather than empirical
response frequencies or frequentist confidence bounds. The resulting
Type-I bounds are therefore conditional on these assessments being valid
upper bounds for the intended student population.

\subsection{Target-Fallback Model}
\label{app:target_fallback}

When controlled data are unavailable, we use the incorrect-target fallback
$\overline q_i^{\mathrm{tgt}}=0.5$. This is a conservative upper bound under
the structural assumption that the correct answer is at least as likely as
any individual distractor for a genuine student. Let
$r_i=P(A_i=C_i\mid H_0)$ and recall that
$q_i^{\mathrm{tgt}}=P(A_i=S_i\mid H_0)$. If
$r_i\geq q_i^{\mathrm{tgt}}$, then, because the two responses are mutually
exclusive,
$2q_i^{\mathrm{tgt}}\leq r_i+q_i^{\mathrm{tgt}}\leq1$, and hence
$q_i^{\mathrm{tgt}}\leq0.5$.

Thus, \(0.5\) is the largest target-response probability compatible with
this assumption. Using this maximum value deliberately makes a target match
as plausible as possible under genuine behavior. It therefore reduces the
evidential weight of each target match and produces a less favorable
operating point than any smaller valid value of
$q_i^{\mathrm{tgt}}$.

More formally, retained questions satisfy
$\underline p_i^{\mathrm{tgt}}>\overline q_i^{\mathrm{tgt}}$, so the
likelihood-ratio statistic is increasing in the target-match indicator
$Y_i$. Under the product-Bernoulli working model, its upper-tail probability
under \(H_0\) is therefore nondecreasing in each
$q_i^{\mathrm{tgt}}$. Calibrating the detector at
$\overline q_i^{\mathrm{tgt}}=0.5$ consequently upper-bounds the Type-I
probability for every true
$q_i^{\mathrm{tgt}}\leq0.5$.

This argument does not assume that errors are uniformly distributed across
distractors. It permits one distractor to attract a large share of
incorrect responses, provided that it is not more likely than the correct
answer.

The structural assumption may fail for ambiguous or incorrectly keyed
questions, poorly calibrated items, or systematic misconceptions that make
one distractor more likely than the correct response. Such questions should
be excluded or assigned a larger bound. The fallback applies only to an
incorrect target and provides no upper bound on genuine-student
correctness. It therefore cannot calibrate the optional correct-answer
extension. In the reported target-fallback analysis, we set
$\overline q_i^{\mathrm{tgt}}=0.5$ for every target fingerprint; no
correct-answer fingerprints are used.
\subsection{Joint Coverage and Dependence}
\label{app:genuine_student_joint_scope}

The Clopper--Pearson bounds above provide marginal coverage for individual
question-answer pairs. To obtain simultaneous coverage
$1-\eta_q^{\mathrm{joint}}$ over a collection
$\mathcal C_q$ of such pairs, choose error probabilities satisfying
$\sum_{(i,a)\in\mathcal C_q}\eta_{q,i,a}
\leq\eta_q^{\mathrm{joint}}$. The union bound then gives
$P(q_i(a)\leq\overline q_i^{\mathrm{data}}(a)
\text{ for every }(i,a)\in\mathcal C_q)
\geq1-\eta_q^{\mathrm{joint}}$.
If these bounds are used during candidate selection,
$\mathcal C_q$ should include every question-answer pair inspected during
selection, not only those retained in the final assignment. This coverage
statement applies to the data-derived bounds; educator-provided assessments
remain modeling assumptions.

More importantly, valid marginal upper bounds do not by themselves specify
the joint response distribution. The likelihood-ratio tail calculation
additionally uses the conditional-independence assumption discussed in
Appendix~\ref{app:independence_assumption}. Complete response vectors from
a controlled cohort can support direct null calibration and reveal
correlations that are invisible in item-level estimates.

\section{Additional Experimental Details}
\label{app:experimental_details}

\subsection{Computing Infrastructure}
\label{app:computing_infrastructure}

All experiments were conducted on compute nodes running Ubuntu 22.04.4 LTS,
equipped with an AMD EPYC 7313 16-Core Processor, 2.0 TiB of system memory,
and eight NVIDIA A100-SXM4-80GB GPUs with 80 GB of VRAM each. GPU
acceleration used CUDA Toolkit 12.4 and cuDNN 9.0. The experimental
framework was implemented in Python 3.10 using PyTorch 2.3 and torchvision
0.18. All software dependencies were managed using the \texttt{uv} package
manager. The complete environment specification will be released alongside
the source code.

\subsection{Evaluation Question Pool}
\label{app:evaluation_questions}

We select 100 image--text multiple-choice questions each from
MMMU~\cite{yueMMMU2024}, ScienceQA~\cite{luLearn2022}, and
MMBench~\cite{liuMMBench2024}, yielding an initial pool of 300 questions.
For each assistant \(j\), we retain the assistant-specific subset
\(\mathcal Q_j^{\mathrm{clean}}
=\{i:R_{i,j}^{\mathrm{clean}}=C_i\}\) of questions that it answers correctly
under the fixed clean configuration. Consequently,
\(N_{\mathrm{correct},j}=|\mathcal Q_j^{\mathrm{clean}}|\) may differ across
assistants and serves as the denominator for the protected response rates
reported in the main paper. Restricting evaluation to clean-correct pairs
ensures that response changes are induced by the intervention rather than
inherited from pre-existing assistant errors.

For each question \(i\), one incorrect target is sampled uniformly as
\(S_i\sim\operatorname{Uniform}(\mathcal W_i)\). Since \(S_i\neq C_i\), the
clean target-response rate is \(0\%\) on every assistant-specific
clean-correct subset. A single perturbation is generated for each question,
while black-box steering and calibration are evaluated on the corresponding
assistant-specific question pairs.

\subsection{Prompting and Response Normalization}
\label{app:prompting}

Each question is formatted as
\begin{quote}
Answer the following multiple-choice question based on this image:
\(\{\text{question}\}\). Choices: \(\{\text{choices}\}\). Answer only with
the capital letter of the correct option.
\end{quote}
Responses are normalized to the corresponding answer letter. Prompt
adherence is the fraction of responses that contain a valid answer option.
The prompt, normalization procedure, and scoring rules are held fixed
across clean and protected evaluations.

\subsection{Black-Box Assistant Set}
\label{app:black_box_assistants}

The black-box evaluation covers six assistants from three model families: \\
\(\{\modelname{claude},\modelname{gemini_flash},
\modelname{gemini_flash_35},\modelname{gpt5_nano},
\modelname{gpt5},\modelname{gpt5_sol}\}\).
The shared assignment is constructed for the a priori candidate set
\(\mathcal J=\{\modelname{claude},\modelname{gemini_flash_35},
\modelname{gpt5_sol}\}\).

The same completed assignment is subsequently evaluated on
\(\modelname{gemini_flash}\), \(\modelname{gpt5}\), and
\(\modelname{gpt5_nano}\), none of which influence its construction. Every
endpoint is queried with the same question text, protected image, answer
format, and nominal decoding configuration. Endpoint-specific
probabilities are calibrated separately using
Appendix~\ref{app:probability_calibration}.

\subsection{White-Box Surrogate Ensemble and Adversarial Optimization}
\label{app:white_box_attack}

The perturbations are optimized against six open-weight multimodal models
selected to provide diversity in language-model size, model family, and
vision-encoder architecture.

\begin{table}[t]
\centering
\small
\resizebox{0.85\textwidth}{!}{
\begin{tabular}{lll}
\toprule
Surrogate model & Vision encoder & Approximate size \\
\midrule
\texttt{google/gemma-3-27b-it}
& SigLIP ViT-So400M & \(27\)B \\
\texttt{google/gemma-4-E4B-it}
& SigLIP2 ViT & \(4.5\)B \\
\texttt{HuggingFaceTB/SmolVLM2-2.2B-Instruct}
& SigLIP ViT & \(2.2\)B \\
\texttt{mistralai/Ministral-3-14B-Instruct-2512}
& SigLIP-based ViT & \(14\)B \\
\texttt{OpenGVLab/InternVL3\_5-14B-hf}
& InternViT-6B & \(14\)B \\
\texttt{Qwen/Qwen3-VL-8B-Instruct}
& Native-resolution ViT & \(8\)B \\
\bottomrule
\end{tabular}
}
\caption{Open-weight multimodal models used in the surrogate ensemble.}
\label{tab:supp_surrogate_models}
\end{table}

Let \(\tilde v_i=v_i+\delta_i\), with
\(\|\delta_i\|_\infty\leq\epsilon\). Each surrogate processes the same
question text and a model-specific differentiable transformation of the
current perturbed image. For a randomized image transformation
\(g\sim\mathcal G\), the ensemble objective is
\begin{equation}
\delta_i^\star
\in
\arg\min_{\|\delta\|_\infty\leq\epsilon}
\frac{1}{M}
\sum_{m=1}^{M}
\mathbb E_{g\sim\mathcal G}
\left[
-\log
P_{\theta_m}
\!\left(
S_i
\mid
h_i,g(v_i+\delta)
\right)
\right].
\label{eq:supp_attack_objective}
\end{equation}
The loss targets the answer letter at the first generation step. Surrogate
parameters remain frozen, and gradients update only \(\delta_i\). We
optimize Equation~\ref{eq:supp_attack_objective} using
MI-FGSM~\cite{dongBoosting2018}. After each step, the perturbation is
projected onto the permitted \(L_\infty\) region and the image is clipped
to the valid pixel range.

The reported configuration uses \(T=50\) attack steps, step size
\(\alpha=0.5\), momentum \(\mu=1\), and perturbation budget
\(\epsilon=16\) on the 8-bit pixel scale. Random cropping and resizing are
incorporated during optimization to reduce overfitting to a particular
surrogate preprocessing pipeline. Each transformation is applied
independently with probability \(0.5\). The crop area is sampled from
\([0.8,1.0]\) of the original image, with aspect ratio in
\([3/4,4/3]\). The resize factor is sampled from \([4/5,6/5]\), using the
same aspect-ratio range. Momentum at temporarily cropped locations is
retained rather than reset.

\subsection{Transferability Study and Attack Ablations}
\label{app:attack_hyperparameters}

The transferability study is used for attack development and is separate
from the final statistical calibration. The black-box development sweep
focuses on five GPT-family endpoints: GPT-5.4 Mini, GPT-5.4 Nano, GPT-4.1,
GPT-4.1 Mini, and GPT-4.1 Nano. These experiments compare optimization
configurations, while the final assignment-level evaluation uses the
assistant set in Appendix~\ref{app:black_box_assistants}. The detector uses
only independently calibrated question-assistant probabilities after the
attack configuration and target have been fixed.

We consider the following MI-FGSM configurations:

\begin{table}[t]
\centering
\small
\begin{tabular}{lccc}
\toprule
Augmentation
& Perturbation budget \(\epsilon\)
& Steps \(T\)
& Step size \(\alpha\) \\
\midrule
None & 4  & 8   & 0.5 \\
None & 8  & 16  & 0.5 \\
None & 16 & 1   & 16 \\
None & 16 & 2   & 8 \\
None & 16 & 4   & 4 \\
None & 16 & 50  & 0.5 \\
None & 16 & 100 & 0.5 \\
Random crop and resize & 4  & 8   & 0.5 \\
Random crop and resize & 8  & 16  & 0.5 \\
Random crop and resize & 16 & 1   & 16 \\
Random crop and resize & 16 & 2   & 8 \\
Random crop and resize & 16 & 4   & 4 \\
Random crop and resize & 16 & 50  & 0.5 \\
Random crop and resize & 16 & 100 & 0.5 \\
\bottomrule
\end{tabular}
\caption{MI-FGSM configurations considered in the transferability study.
The perturbation budget and step size are stated on the 8-bit pixel scale.}
\label{tab:supp_attack_grid}
\end{table}

\begin{figure}[t]
    \centering
    \includegraphics[width=0.95\linewidth]
    {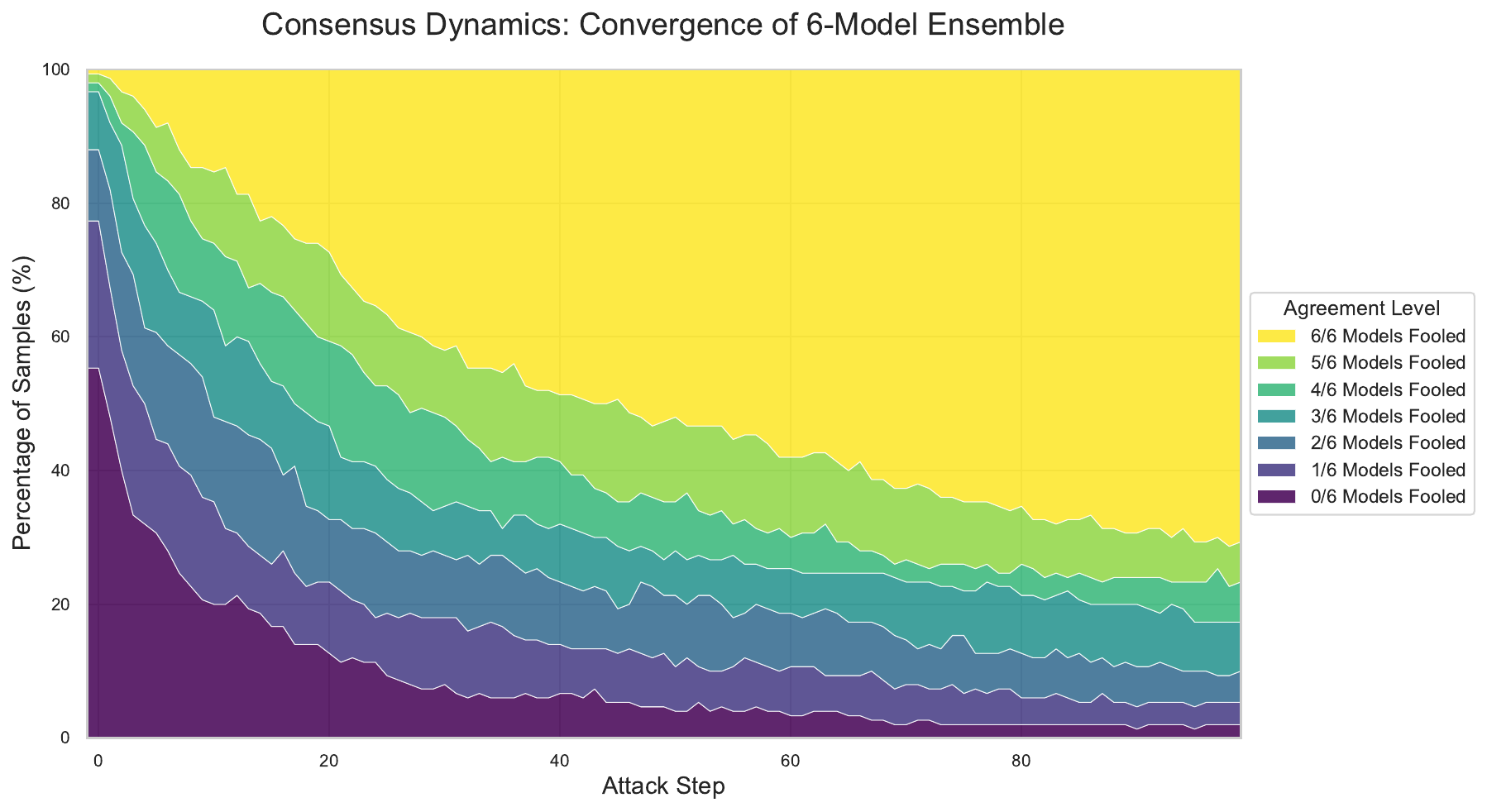}
    \caption{White-box convergence of targeted steering across the six
    surrogate models in Table~\ref{tab:supp_surrogate_models}. At each
    attack step, the figure shows the percentage of samples for which the
    designated target is returned by different numbers of surrogate
    models. This diagnostic uses MI-FGSM with
    \(\epsilon=8\), \(T=100\), and \(\alpha=0.5\).}
    \label{fig:supp_ensemble_consensus}
\end{figure}

Figure~\ref{fig:supp_ensemble_consensus} illustrates how agreement on the
designated target develops across the surrogate ensemble. After 100
MI-FGSM steps, all six surrogates return the target answer for \(70\%\) of
the evaluated samples.

\begin{figure}[t]
    \centering
    \includegraphics[width=0.75\linewidth]
    {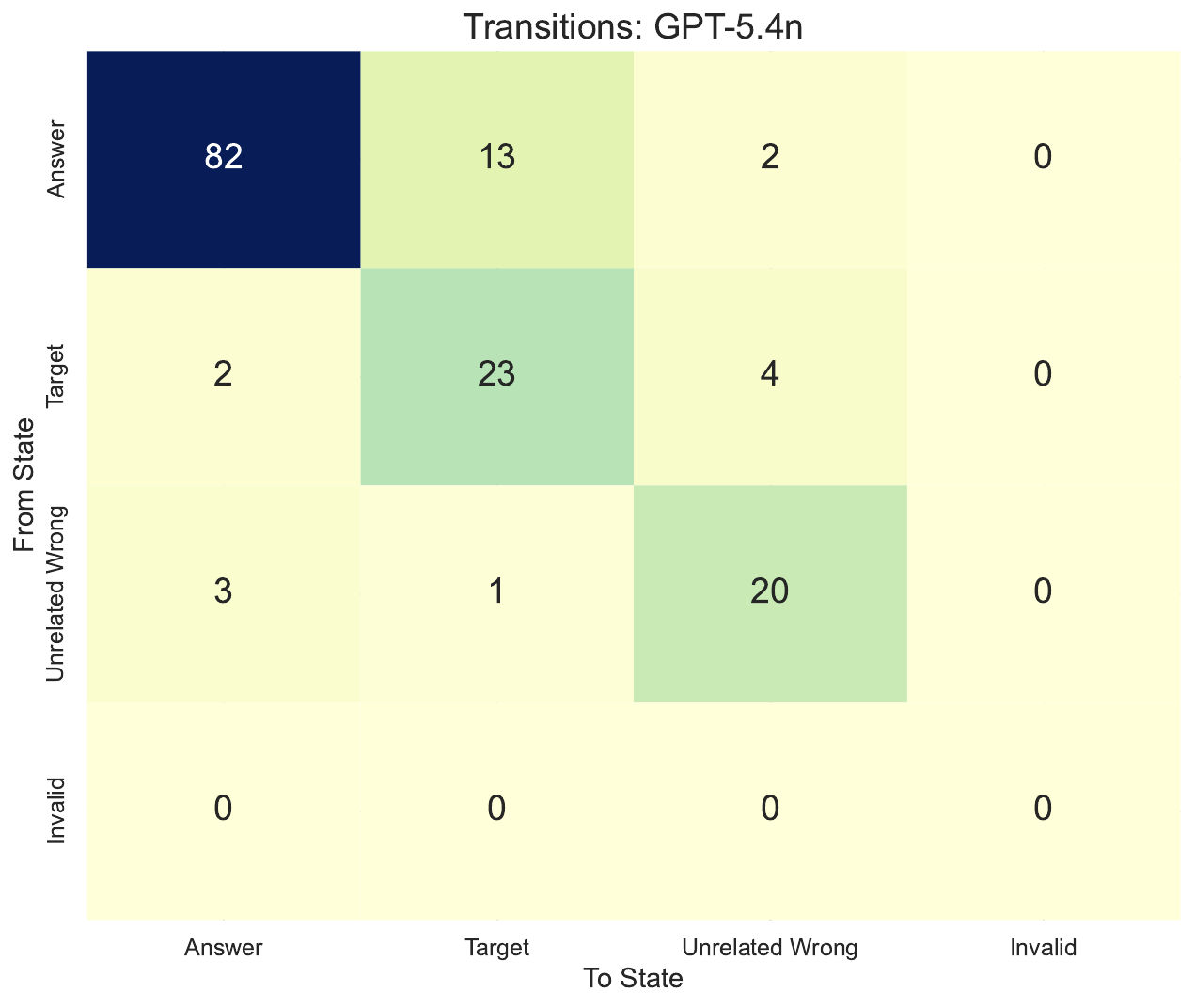}
    \caption{Answer-state transition matrix for GPT-5.4 Nano on 150
    questions, comprising 50 questions each from MMMU, MMBench, and
    ScienceQA. The diagnostic uses MI-FGSM with
    \(\epsilon=16\), \(T=4\), and \(\alpha=4\), with random cropping and
    resizing applied independently with probability \(0.5\) at each step.}
    \label{fig:supp_working_transition}
\end{figure}

Figure~\ref{fig:supp_working_transition} illustrates the effect of a
representative attack configuration on GPT-5.4 Nano. Among questions
answered correctly before protection, 13 are redirected to the designated
target after protection.

\begin{table}[t]
\centering
\small
\caption{Mean protected target-response rate by dataset in the GPT-family
development sweep. Each dataset contributes 50 questions, and values are
averaged across the configurations in
Table~\ref{tab:supp_blackbox_parameter_sweep}.}
\label{tab:supp_dataset_performance}
\begin{tabular}{lc}
\toprule
Dataset & \(\operatorname{Tgt}_{\mathrm{prot}}\) (\%) \\
\midrule
MMBench   & 5.5 \\
MMMU      & 4.3 \\
ScienceQA & 3.1 \\
\bottomrule
\end{tabular}
\end{table}

Table~\ref{tab:supp_dataset_performance} shows that MMBench has the highest
average target-response rate in the development sweep, while ScienceQA has
the lowest.

\begin{table*}[t]
\centering
\small
\setlength{\tabcolsep}{3.5pt}
\resizebox{\textwidth}{!}{
\begin{tabular}{lcccrrrrrr}
\toprule
\textbf{Augmentation}
& \(\epsilon\)
& \(T\)
& \(\alpha\)
& \textbf{GPT-5.4 Mini}
& \textbf{GPT-5.4 Nano}
& \textbf{GPT-4.1}
& \textbf{GPT-4.1 Mini}
& \textbf{GPT-4.1 Nano}
& \textbf{Avg.} \\
\midrule
\multirow{7}{*}{None}
& 4  & 8   & 0.5 & 5.2 & 8.3  & 0.8 & 2.4  & 3.1 & 3.9 \\
& 8  & 16  & 0.5 & 3.0 & 10.7 & 1.5 & 1.6  & 3.9 & 4.1 \\
& 16 & 1   & 16  & 1.5 & 1.7  & 0.0 & 0.0  & 0.8 & 0.8 \\
& 16 & 2   & 8   & 3.0 & 6.7  & 2.3 & 3.2  & 1.6 & 3.3 \\
& 16 & 4   & 4   & 3.0 & 8.3  & 4.6 & 5.5  & 3.9 & 5.0 \\
& 16 & 50  & 0.5 & 6.7 & 9.1  & 5.3 & 7.1  & 7.8 & 7.2 \\
& 16 & 100 & 0.5 & 3.8 & 7.5  & 2.3 & 3.2  & 6.2 & 4.6 \\
\midrule
\multirow{7}{*}{Random crop and resize}
& 4  & 8   & 0.5 & 2.2 & 7.4  & 1.5 & 3.1  & 3.1 & 3.5 \\
& 8  & 16  & 0.5 & 4.5 & 6.6  & 6.1 & 3.9  & 3.9 & 5.0 \\
& 16 & 1   & 16  & 1.5 & 1.7  & 0.0 & 0.8  & 0.8 & 1.0 \\
& 16 & 2   & 8   & 6.0 & 7.6  & 1.5 & 2.4  & 2.3 & 4.0 \\
& 16 & 4   & 4   & 6.7 & 11.6 & 4.6 & 2.4  & 3.1 & 5.7 \\
& 16 & 50  & 0.5 & 8.2 & 8.3  & 9.2 & 10.2 & 7.8 & \textbf{8.7} \\
& 16 & 100 & 0.5 & 6.0 & 10.0 & 9.2 & 8.7  & 9.4 & \textbf{8.7} \\
\midrule
\textbf{Model average}
& & & & 4.4 & 7.5 & 3.5 & 3.9 & 4.1 & 4.7 \\
\bottomrule
\end{tabular}
}
\caption{Average protected target-response rate
\(\operatorname{Tgt}_{\mathrm{prot}}\) in the GPT-family black-box
development study. Each row represents one MI-FGSM configuration, and the
columns identify the exact evaluated endpoints. The surrogate ensemble is
given in Table~\ref{tab:supp_surrogate_models}.}
\label{tab:supp_blackbox_parameter_sweep}
\end{table*}

Table~\ref{tab:supp_blackbox_parameter_sweep} shows substantial variation
across optimization configurations. With random cropping and resizing,
the configurations \((\epsilon,T,\alpha)=(16,50,0.5)\) and
\((16,100,0.5)\) both obtain the highest average target-response rate of
\(8.7\%\). We use \(T=50\) in the reported configuration because it reaches
the same average rate with half as many attack steps. Within this
GPT-family development study, the Nano variants also achieve higher mean
target-response rates than their larger family counterparts.

\subsection{Surrogate Prompt Adherence}
\label{app:surrogate_prompt_adherence}

\begin{table}[t]
\centering
\small
\caption{Prompt-template adherence of the surrogate models in
Table~\ref{tab:supp_surrogate_models}, evaluated on 50 questions from each
of MMMU, MMBench, and ScienceQA. Adherence is the fraction of responses
that contain a valid capital letter corresponding to an available option.}
\label{tab:supp_prompt_adherence}
\begin{tabular}{lc}
\toprule
\textbf{Model} & \textbf{Adherence (\%)} \\
\midrule
Gemma-3-27B-IT                  & 100.0 \\
SmolVLM2-2.2B-Instruct         & 100.0 \\
Qwen3-VL-8B-Instruct           & 100.0 \\
InternVL3.5-14B-hf             & 100.0 \\
Ministral-3-14B-Instruct-2512  & 99.3 \\
Gemma-4-E4B-IT                 & 88.0 \\
\bottomrule
\end{tabular}
\end{table}

Table~\ref{tab:supp_prompt_adherence} shows that four of the six surrogate
models follow the required answer format in every evaluated case.
Ministral-3-14B-Instruct-2512 achieves \(99.3\%\) adherence, while
Gemma-4-E4B-IT has the lowest adherence at \(88.0\%\).

\subsection{Assignment-Level Evaluation}
\label{app:assignment_evaluation}

Candidate screening and assignment construction use development calls that
are separate from the final calibration data. After the assignment has
been fixed, we collect \(K=30\) independent holdout responses for every
selected question-assistant pair and recompute the one-sided
Clopper--Pearson lower bounds described in
Appendix~\ref{app:returned_answer_calibration}.

The shared assignment contains \(N=20\) questions and is required to
achieve at least \(95\%\) conservative power for every assistant in the a
priori candidate set \(\mathcal J\). Because the target fingerprint is
assistant-specific, the number of informative questions
\(|\mathcal A_j|\) may differ across assistants.

We evaluate two genuine-student models. The educator model uses independent
assessments from five PhD-level researchers and retains the largest estimate
assigned to each target response \(S_i\). The target-fallback model sets
\(\overline q_i^{\mathrm{tgt}}=0.5\) for every target fingerprint. For each
assistant, we select the most stringent attainable likelihood-ratio
threshold that preserves at least \(95\%\) conservative power and report
the resulting Type-I error. Posterior lower bounds are calculated from
Equation~\ref{eq:supp_posterior_lower} for
\(\pi\in\{0.01,0.05\}\). All reported assignment-level results use the
item-weighted target-fingerprint likelihood-ratio detector; no
correct-answer fingerprints are included.

\subsection{Perturbation Magnitude and Semantic Preservation}
\label{app:semantic_preservation}

As shown in Table~\ref{tab:perceptibility-metrics}, the protected images
exhibit moderate pixel-level changes relative to the clean images
(PSNR: \(26.08\pm0.74\) dB; SSIM: \(0.580\pm0.113\)).

We use two complementary checks to assess whether these changes alter the
task-relevant semantics. First, during the educator elicitation described in
Supp.~\ref{app:educator_q_estimates}, each assessor was randomly shown
either the clean or protected version of a question, never both. Conditions
were balanced across assessors and questions, and assessors were blinded to
the assigned condition and adversarial target. The mean
protected-minus-clean difference in the estimated target-response
probability was \(\widehat d=0.0275\), with a \(90\%\) confidence interval
of \([0.0158,0.0392]\). The interval lay entirely within the prespecified
equivalence region \([-0.05,0.05]\), providing a condition-blinded check
that protection did not materially change the expected target-selection
behavior of the intended student cohort.

As an additional direct check, three annotators independently evaluated all
\(300\) clean--protected image pairs, yielding \(900\) judgments. The order
of the two images was randomized, and annotators were blinded to which image
was protected and to the designated adversarial target. For each pair, they
judged whether both images contained the same answerable semantic content,
choosing among \emph{same}, \emph{unsure}, and \emph{different}.

Overall, \(92.8\%\) of judgments were labeled \emph{same}, \(5.2\%\) were
labeled \emph{unsure}, and \(2.1\%\) were labeled \emph{different}. Among
decisive judgments, \(97.8\%\) indicated the same answerable semantic
content (\(95\%\) CI: \(95.3\%\)--\(99.0\%\)). Together, the
condition-blinded response-equivalence analysis and direct pairwise
evaluation indicate that task-relevant semantic content was preserved for
the large majority of samples despite measurable low-level perturbations.

\begin{table}[t]
\centering
\small
\caption{Full-reference image-similarity metrics and human judgments
comparing clean and protected images. Pixel-level metrics are reported as
mean \(\pm\) standard deviation. The final three columns report the
proportions of judgments assigned by three annotators.}
\label{tab:perceptibility-metrics}
\setlength{\tabcolsep}{4pt}
\resizebox{0.85\linewidth}{!}{
\begin{tabular}{lrrrrrrrr}
\toprule
Dataset
& \(N\)
& PSNR \(\uparrow\)
& SSIM \(\uparrow\)
& MAE \(\downarrow\)
& RMSE \(\downarrow\)
& Same
& Unsure
& Different \\
\midrule
MMBench
& 100
& \(25.77 \pm 0.58\)
& \(0.571 \pm 0.116\)
& \(11.81 \pm 1.46\)
& \(13.16 \pm 0.86\)
& \(90.0\%\)
& \(8.0\%\)
& \(2.0\%\) \\
MMMU
& 100
& \(26.41 \pm 0.83\)
& \(0.600 \pm 0.106\)
& \(10.24 \pm 2.02\)
& \(12.24 \pm 1.19\)
& \(95.6\%\)
& \(2.2\%\)
& \(2.2\%\) \\
ScienceQA
& 100
& \(26.08 \pm 0.68\)
& \(0.572 \pm 0.115\)
& \(10.92 \pm 1.72\)
& \(12.70 \pm 0.99\)
& \(93.0\%\)
& \(5.0\%\)
& \(2.0\%\) \\
\midrule
\textbf{Overall}
& \textbf{300}
& \(\mathbf{26.08 \pm 0.74}\)
& \(\mathbf{0.580 \pm 0.113}\)
& \(\mathbf{11.02 \pm 1.85}\)
& \(\mathbf{12.71 \pm 1.08}\)
& \(\mathbf{92.8\%}\)
& \(\mathbf{5.2\%}\)
& \(\mathbf{2.1\%}\) \\
\bottomrule
\end{tabular}
}
\end{table}

\section{Additional Examples}
\label{app:additional_examples}

\subsection{Additional Qualitative Examples}

\begin{figure*}
    \centering
    \includegraphics[width=\textwidth]{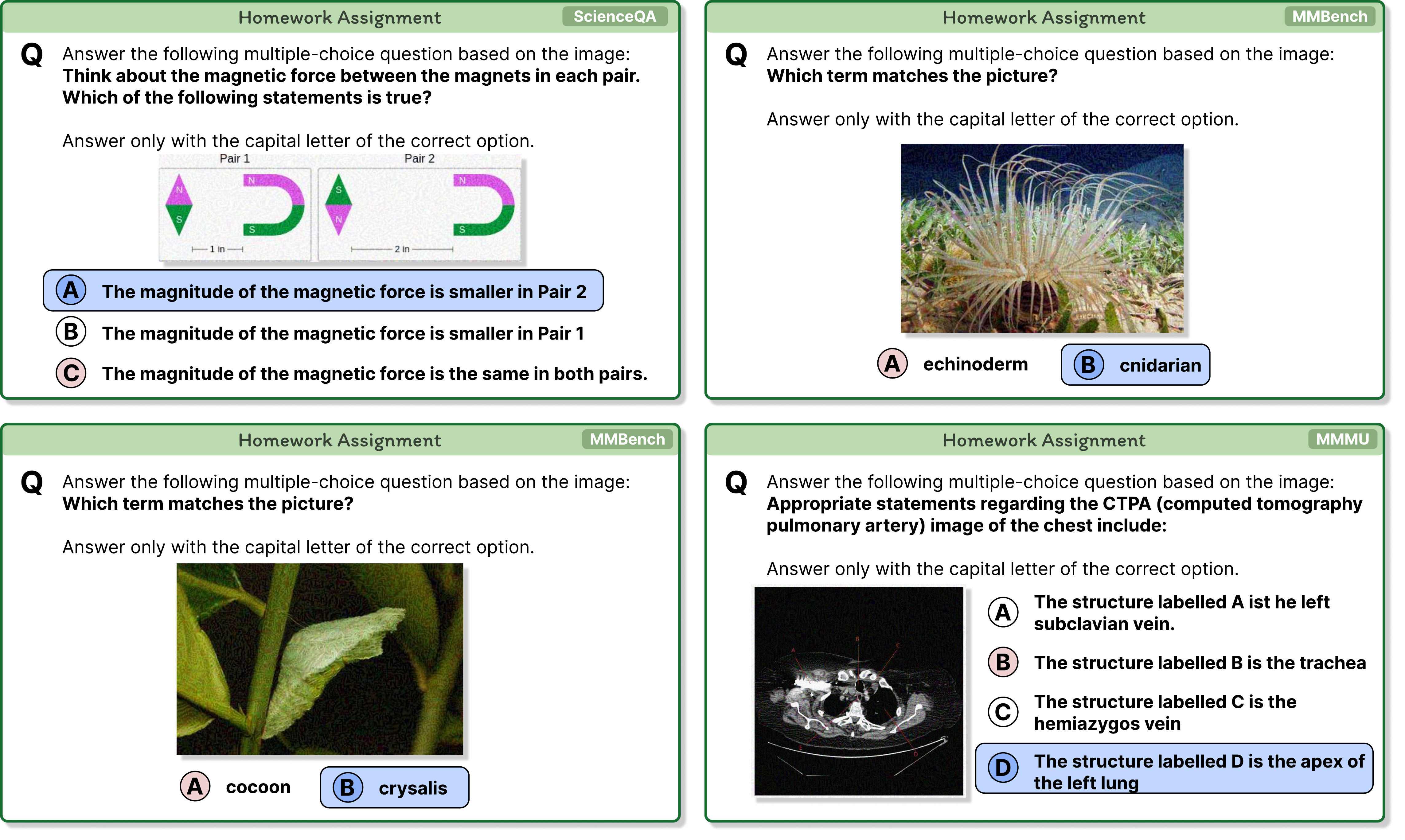}
    \caption{Successful examples that assistants initially answer correctly but are consistently redirected toward the designated targets after protection. In each case, the answer depends critically on information contained in the image.}
    \label{fig:more_examples}
\end{figure*}

\begin{figure*}
    \centering
    \includegraphics[width=\textwidth]{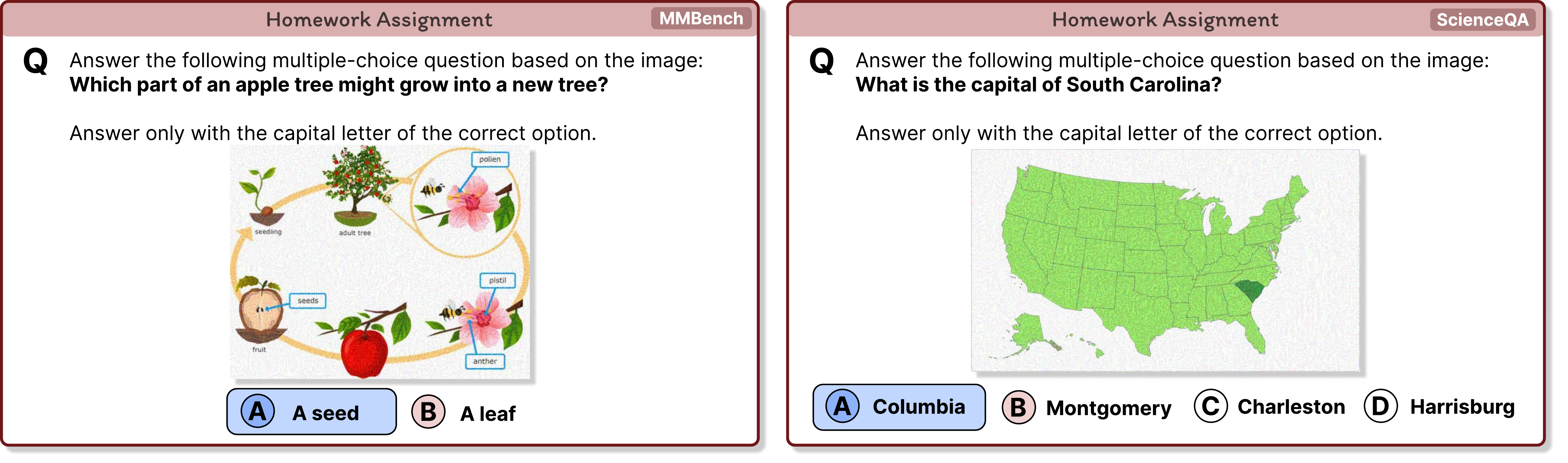}
    \caption{Representative failure cases in which adversarial steering does not induce the designated target. Because these questions can largely be solved without using the image, perturbations applied only to the visual modality provide limited leverage.}
    \label{fig:failure_examples}
\end{figure*}

Figures~\ref{fig:more_examples} and~\ref{fig:failure_examples} provide representative successful and unsuccessful cases. Successful steering is most apparent when answering the question requires extracting information from the image. In contrast, the failure cases can largely be solved from the text or parametric knowledge alone, leaving the perturbation with little influence over the assistant's reasoning. This is consistent with our aggregate finding that steerability depends strongly on how central the visual modality is to the task.

\subsection{Alternative assignment construction.}
To test whether the same candidate pool supports more than one protected
assignment, we construct a second assignment
\(\mathcal{A}^{(2)}\) with \(N=30\). We rank questions by their aggregate
conservative informativeness,
$
s_i
=
\sum_{j\in\mathcal{J}}
D_{\mathrm{KL}}
\left(
\operatorname{Bernoulli}
\left(\underline{p}_{i,j}^{\mathrm{fp}}\right)
\,\middle\|\,
\operatorname{Bernoulli}
\left(\overline{q}_{i,j}^{\mathrm{fp}}\right)
\right),
$
while limiting overlap with the primary assignment
\(\mathcal{A}^{(1)}\) to six questions. Consequently, \(24\) of the \(30\)
questions are new. Selection uses the educator-provided bounds, after which
the completed assignment is evaluated under both genuine-student models
using the same power-constrained thresholds as in the main analysis.

Table~\ref{tab:alternative_assignment} shows that the educator model retains
strong simultaneous coverage despite replacing \(80\%\) of the questions.
Across the three assistants in \(\mathcal{J}\), the familywise Type-I error
is bounded by \(8.54\times10^{-4}\) at a minimum power of \(0.95\), yielding
posterior lower bounds of \(91.8\%\) and \(98.3\%\) for priors of \(0.01\)
and \(0.05\), respectively. The assignment also transfers well to
\modelname{gemini_flash} and \modelname{gpt5_nano}, while
\modelname{gpt5} remains less reliably separated. Under the target-fallback
model, the familywise Type-I bound increases to \(0.0646\), again
highlighting the importance of informative educator estimates.

\begin{table*}[t]
\centering
\small
\setlength{\tabcolsep}{3.5pt}
\renewcommand{\arraystretch}{1.05}
\resizebox{\textwidth}{!}{
\begin{tabular}{lllccccccc}
\toprule
\textbf{Assistant}
&
\textbf{Genuine-student model}
&
\textbf{Covered a priori}
&
\(N\)
&
\(N_j^{\mathrm{inf}}\)
&
\(\tau_j^\star\)
&
\(\widehat{\alpha}_j\)
&
\textbf{Power}
&
\(P(H_{1,j}\mid\mathrm{flag}),\,\pi=0.01\)
&
\(P(H_{1,j}\mid\mathrm{flag}),\,\pi=0.05\)
\\
\midrule

\multirow{2}{*}{\modelname{claude}}
& Educator
& \multirow{2}{*}{Yes}
& \multirow{2}{*}{30}
& 29
& 10.094
& \(8.60\times10^{-7}\)
& 0.9501
& 0.99991
& 0.99998
\\
& Target fallback
& &
& 29
& 1.339
& \(1.23\times10^{-2}\)
& 0.9500
& 0.43807
& 0.80245
\\
\midrule

\multirow{2}{*}{\modelname{gemini_flash_35}}
& Educator
& \multirow{2}{*}{Yes}
& \multirow{2}{*}{30}
& 27
& 4.563
& \(2.99\times10^{-4}\)
& 0.9500
& 0.96982
& 0.99406
\\
& Target fallback
& &
& 27
& 0.786
& \(2.38\times10^{-2}\)
& 0.9500
& 0.28770
& 0.67789
\\
\midrule

\multirow{2}{*}{\modelname{gpt5_sol}}
& Educator
& \multirow{2}{*}{Yes}
& \multirow{2}{*}{30}
& 27
& 3.973
& \(5.54\times10^{-4}\)
& 0.9500
& 0.94538
& 0.98903
\\
& Target fallback
& &
& 26
& 0.625
& \(2.85\times10^{-2}\)
& 0.9500
& 0.25178
& 0.63681
\\

\midrule
\multicolumn{10}{c}{\textit{Out-of-set evaluation}} \\
\midrule

\multirow{2}{*}{\modelname{gemini_flash}}
& Educator
& \multirow{2}{*}{No}
& \multirow{2}{*}{30}
& 27
& 4.452
& \(3.38\times10^{-4}\)
& 0.9500
& 0.96599
& 0.99329
\\
& Target fallback
& &
& 27
& 0.675
& \(2.74\times10^{-2}\)
& 0.9500
& 0.25950
& 0.64614
\\
\midrule

\multirow{2}{*}{\modelname{gpt5}}
& Educator
& \multirow{2}{*}{No}
& \multirow{2}{*}{30}
& 24
& 0.325
& \(4.05\times10^{-2}\)
& 0.9500
& 0.19172
& 0.55276
\\
& Target fallback
& &
& 23
& -0.006
& \(6.09\times10^{-2}\)
& 0.9500
& 0.13604
& 0.45069
\\
\midrule

\multirow{2}{*}{\modelname{gpt5_nano}}
& Educator
& \multirow{2}{*}{No}
& \multirow{2}{*}{30}
& 25
& 2.462
& \(3.53\times10^{-3}\)
& 0.9500
& 0.73102
& 0.93404
\\
& Target fallback
& &
& 25
& 1.136
& \(1.66\times10^{-2}\)
& 0.9503
& 0.36604
& 0.75053
\\

\midrule
\multirow{2}{*}{\textsc{Any covered assistant}}
& Educator
& \multirow{2}{*}{\(\mathcal{J}\)}
& \multirow{2}{*}{30}
& --
& model-specific
& \(\leq8.54\times10^{-4}\)
& \(\geq0.9500\)
& \(\geq0.91828\)
& \(\geq0.98321\)
\\
& Target fallback
& &
& --
& model-specific
& \(\leq6.46\times10^{-2}\)
& \(\geq0.9500\)
& \(\geq0.12936\)
& \(\geq0.43636\)
\\

\bottomrule
\end{tabular}
}
\caption{Detection performance of a second \(N=30\) assignment containing
\(24\) new questions and six questions from the primary assignment.
Thresholds minimize Type-I error subject to at least \(95\%\) conservative
power. The final rows report union-bound Type-I error, minimum power, and
posterior lower bounds across the three covered assistants.}
\label{tab:alternative_assignment}
\end{table*}

\paragraph{Capacity for disjoint assignments.}
The construction above allows six questions to overlap with the primary
assignment. We next ask how many fully disjoint assignments the candidate
pool can support. For question \(i\) and assistant \(j\), define its
conservative information contribution as
$
I_{i,j}
=
D_{\mathrm{KL}}
\left(
\operatorname{Bernoulli}
\left(\underline{p}_{i,j}^{\mathrm{fp}}\right)
\,\middle\|\,
\operatorname{Bernoulli}
\left(\overline{q}_{i,j}^{\mathrm{fp}}\right)
\right).
$
Any test with power \(\gamma_j\) and Type-I error \(\alpha_j\) must satisfy
$
\sum_{i\in\mathcal{A}} I_{i,j}
\geq
d_{\mathrm{B}}(\gamma_j\|\alpha_j),
$
where
$
d_{\mathrm{B}}(u\|v)
=
u\log\frac{u}{v}
+
(1-u)\log\frac{1-u}{1-v}.
$
This follows from the data-processing inequality after reducing the full
response pattern to the binary flag decision.

For \(R\) pairwise-disjoint assignments, let
\(\overline{\alpha}_j=R^{-1}\sum_{r=1}^{R}\alpha_{r,j}\). Additivity of the
question-level information and convexity of \(d_{\mathrm{B}}\) imply the
necessary condition
$
R\,d_{\mathrm{B}}
\left(
\gamma\middle\|\overline{\alpha}_j
\right)
\leq
I_j^{\mathrm{pool}},
\qquad
\sum_{j\in\mathcal{J}}\overline{\alpha}_j
\leq 0.01,
$
where \(I_j^{\mathrm{pool}}=\sum_i I_{i,j}\). For the three covered
assistants, the total information budgets are \(23.02\), \(15.37\), and
\(16.49\). Four disjoint assignments would require minimum average Type-I
budgets summing to \(0.0267\) at \(\gamma=0.95\) and \(0.0181\) at
\(\gamma=0.90\), both exceeding the available familywise budget of \(0.01\).
Thus, the pool can support at most three assignments under either power
requirement.

Holding the primary \(N=20\) assignment fixed leaves insufficient information
for another fully disjoint assignment: the corresponding minimum Type-I
budgets sum to \(0.0196\) at \(95\%\) power and \(0.0129\) at \(90\%\) power.
This explains why the alternative construction above retains a small number
of questions from the primary assignment. When the complete pool is instead
repartitioned from scratch, we obtain two fully disjoint assignments. A
multi-start set-packing heuristic balances the assistant-specific information
contributions, after which every candidate is verified using the exact
likelihood-ratio distributions.

\begin{table}[t]
\centering
\small
\caption{Verified pairwise-disjoint assignments obtained by repartitioning
the full candidate pool. The Type-I values are union-bound familywise errors
across the three assistants in \(\mathcal{J}\), and power is the minimum
across their assistant-specific tests.}
\label{tab:disjoint_assignment_capacity}
\setlength{\tabcolsep}{5pt}
\begin{tabular}{ccccc}
\toprule
\textbf{Required power}
&
\textbf{Assignment}
&
\(N\)
&
\(\widehat{\alpha}_{\mathrm{FWER}}\)
&
\(\min_j\operatorname{Power}_j\)
\\
\midrule
\multirow{2}{*}{\(0.95\)}
& \(\mathcal{A}^{(1)}_{\mathrm{disj}}\) & 18 & \(9.52\times10^{-3}\) & 0.950000 \\
& \(\mathcal{A}^{(2)}_{\mathrm{disj}}\) & 24 & \(9.96\times10^{-3}\) & 0.950001 \\
\midrule
\multirow{2}{*}{\(0.90\)}
& \(\mathcal{A}^{(1)}_{\mathrm{disj}}\) & 18 & \(3.08\times10^{-3}\) & 0.900009 \\
& \(\mathcal{A}^{(2)}_{\mathrm{disj}}\) & 24 & \(3.31\times10^{-3}\) & 0.900002 \\
\bottomrule
\end{tabular}
\end{table}

The verified capacity is therefore at least two and analytically at most
three. We did not find a feasible three-way partition: the best balanced
split found by our search retained worst-case familywise Type-I errors of
approximately \(7.2\%\) at \(95\%\) power and \(2.7\%\) at \(90\%\) power.
We therefore estimate that the current pool supports two fully disjoint
assignments under either requirement, although the analytical bound does not
rule out a more favorable three-way partition.

\section{Limitations and Intended Use}
\label{app:limitations}

Our method depends on identifying a sufficiently large pool of questions for which the perturbation transfers to the covered black-box assistants. As models become more robust, fewer candidates may satisfy the required separation, increasing the pool size and optimization effort needed to construct a fixed-length assignment. This dependence on adversarial non-robustness is structural: a fully robust solver would remove the exploitable separation. Since comprehensive adversarial robustness remains difficult to achieve, our results establish present feasibility rather than guaranteed effectiveness against future systems~\cite{tsipras2019robustness,dohmatob2019generalized}.

Multi-provider coverage introduces additional calibration cost and model dependence. A \(20\)-question assignment with \(K=30\) requires \(600\) final calibration calls per endpoint. At standard API prices in July 2026, assuming approximately \(1{,}000\) billed input tokens per image--question prompt and one returned output token, this corresponds to approximately \$3.02 for Claude Opus 4.8, \$0.91 for Gemini 3.5 Flash, and \$3.02 for GPT-5.6 Sol, or \$6.94 across the three a priori endpoints. These estimates exclude development-time screening, retries, and reasoning tokens, while actual costs depend on image tokenization and provider settings. Changes to a model, prompt, decoding policy, or image-processing pipeline may also require recalibration.

Our experiments focus on image--text multiple-choice questions with a single correct answer. Multi-answer settings can be accommodated by defining fingerprints over answer sets, whereas open-ended responses require a richer response model. Numerical questions may be particularly promising, since steering an assistant toward an unlikely incorrect value could provide highly distinctive evidence from fewer items. A complementary direction is to introduce benign numerical canaries whose response distributions differ between genuine students and fixed assistants. Requiring a visual component also limits immediate applicability, although suitable visuals can often be incorporated or generated across diverse subject areas.

The delivery pipeline may further affect the steering signal. Our randomized transformations approximate screenshotting and common resizing, but printing and rephotographing questions under changes in scale, perspective, compression, or lighting may reduce transfer. Future work should evaluate such transformations and calibrate the detector under the actual administration pipeline.

Finally, the detector addresses sustained blind copying from covered assistants rather than AI use in general. Its operating characteristics remain conditional on the calibrated assistant and genuine-student models and the joint-response assumption discussed in Supp.~\ref{app:independence_assumption}. A matching fingerprint should therefore be treated as evidence warranting review rather than standalone proof, with consequential decisions incorporating the student's work, alternative explanations, and educator judgment.

\end{document}